\documentclass[pdflatex,sn-vancouver-ay]{sn-jnl}% Vancouver Author Year Reference Style
\usepackage{graphicx}%
\usepackage{amsmath,amssymb,amsfonts}%
\usepackage{amsthm}%
\usepackage{xcolor}%
\usepackage{manyfoot}%
\usepackage{tikz}
\usetikzlibrary{arrows.meta,positioning}
\usepackage{subcaption}
\usepackage{algorithm}
\usepackage{algpseudocode}

\usepackage{booktabs,multirow,makecell,array}

\usepackage{multirow}
\usepackage{textcomp}
\usepackage{xcolor}
\usepackage{flushend}
\usepackage{enumitem}

\begin{document}

\title[Article Title]{Stable Global Weighting of Flow Mixtures using Simplex Exponential Moving Average}

\author*[1]{\fnm{Benjamin} \sur{Wiriyapong}}\email{wiriyapongb@cardiff.ac.uk}

\author[1]{\fnm{Oktay} \sur{Karakuş}}\email{karakuso@cardiff.ac.uk}

\author[2,1]{\fnm{Can} \sur{Eyupoglu}}\email{eyupogluc@cardiff.ac.uk}

\author[1]{\fnm{Kirill} \sur{Sidorov}}\email{sidorovk@cardiff.ac.uk}

\affil*[1]{\orgdiv{School of Computer Science and Informatics}, \orgname{Cardiff University}, \orgaddress{\street{Abacws}, \city{Cardiff}, \postcode{CF24 4AG}, \country{U.K.}}}

\affil[2]{\orgdiv{Department of Computer Engineering}, \orgname{Turkish Air Force Academy}, \orgaddress{National Defence University}, \city{Istanbul}, \country{Türkiye}}

%\affil[2]{\orgdiv{Department}, \orgname{Organization}, \orgaddress{\street{Street}, \city{City}, \postcode{10587}, \state{State}, \country{Country}}}

%\affil[3]{\orgdiv{Department}, \orgname{Organization}, \orgaddress{\street{Street}, \city{City}, \postcode{610101}, \state{State}, \country{Country}}}


\abstract{Normalising flows provide a powerful variational family for approximate inference, yet individual architectures often fail to generalise across heterogeneous posterior geometries.
We revisit mixture-based flow formulations and introduce \emph{AMF\mbox{-}VI\mbox{-}sEMA}, a two-stage framework featuring a \emph{stable global weighting} mechanism based on a \emph{Simplex Exponential Moving Average} (sEMA) update.
In Stage~1, a heterogeneous set of experts (\textsc{RealNVP}, \textsc{MAF}, \textsc{RBIG}) are trained independently to specialise in distinct structural regimes.
In Stage~2, expert parameters are frozen and global mixture weights are learned through a temperature-controlled softmax of average log-likelihoods, followed by a smooth EMA update on the probability simplex.
This design produces a tractable, data-agnostic gating mechanism (without per-sample gating or gradient backpropagation through weights) that adaptively reallocates capacity while avoiding component collapse.
We evaluate the framework on ten posterior benchmarks: six canonical 2D synthetic families (Banana, X-Shaped, Bimodal, Multimodal, Two-moons, Rings) and four real/low-dimensional Bayesian targets (BLR, BPR, Weibull, Real-GMM2), with stronger baselines (\textsc{NICE}, \textsc{ResFlow}, and EM-Mixing).
Comprehensive evaluation covers NLL, KL divergence, Wasserstein-2 distance, and MMD, together with diagnostics of mixture dynamics, hyperparameter sensitivity, and cross-seed robustness.
Empirically, \emph{AMF\mbox{-}VI\mbox{-}sEMA} achieves consistent NLL improvements over its predecessor \emph{AMF\mbox{-}VI} and avoids the catastrophic transport failures of single-flow baselines, while maintaining stable weight trajectories ($N_{\mathrm{eff}}{>}1.4$ on all datasets) with minimal computational overhead.
Code, configurations, and checkpoints are publicly released at \url{https://github.com/benjaminsw/AMF-VIJ} to ensure full reproducibility.}


\maketitle

\section{Introduction}

Variational inference (VI) has become one of the principal approaches for scalable Bayesian inference, enabling posterior approximation in probabilistic models that are otherwise computationally intractable \citep{jordan1999introduction,Blei2017VIReview}. By replacing expensive sampling procedures with optimisation over a tractable variational family, VI has become a standard component of modern machine learning, with applications ranging from Bayesian neural networks and probabilistic graphical models to deep generative modelling and scientific inference \citep{bishop2006pattern,kingma2013auto}. Its practical success, however, depends critically on the expressive power of the chosen variational family.

The most widely used variational families remain fundamentally limited when the target posterior exhibits complex multimodal structure. Classical Gaussian approximations are unable to capture multiple modes, skewness, or highly nonlinear dependencies, typically resulting in underestimated posterior uncertainty and biased inference \citep{bishop2006pattern,Blei2017VIReview}. Such multimodal posteriors naturally arise across numerous application domains, including hierarchical Bayesian models, latent variable models, epidemiological inference, financial regime modelling, and inverse problems, where accurately representing multiple plausible solutions is essential for reliable uncertainty quantification rather than merely improving predictive performance \citep{wakefield2007disease,ang2002regime,hamilton1989new}.

Normalising flows have significantly expanded the flexibility of variational inference by transforming simple base distributions into expressive probability densities through sequences of invertible mappings \citep{rezende2015variational,dinh2016realnvp,papamakarios2017masked,papamakarios2021normalizing}. Modern flow architectures such as RealNVP, Masked Autoregressive Flow (MAF), and Rotation-based Iterative Gaussianisation (RBIG) each provide distinct advantages in modelling complex distributions. Nevertheless, no single architecture performs consistently across the wide variety of posterior geometries encountered in practice. Different flows exhibit different inductive biases, optimisation dynamics, and numerical characteristics, often specialising to particular distributional structures while performing less favourably on others. Consequently, selecting an appropriate flow architecture remains highly problem dependent.

A natural strategy is therefore to combine multiple complementary flows within a mixture model. Mixture-based approaches increase representational capacity and have demonstrated improved flexibility over individual components \citep{PiresFigueiredo2020VMoNF,NgZammitMangion2024SphereMixFlow}. However, existing mixture-of-flow frameworks typically rely on one of two strategies. Either mixture components and mixture weights are learned jointly through a single optimisation objective, or globally fixed mixture coefficients are estimated during training and subsequently held constant during inference. Both strategies introduce important limitations. Joint optimisation couples heterogeneous optimisation dynamics across fundamentally different flow architectures, often leading to unstable training or expert domination. Conversely, fixed global weighting lacks the flexibility to exploit complementary expert behaviour during optimisation and frequently allocates representational capacity inefficiently across complex posterior geometries. As a result, increasing the number or diversity of flow architectures alone does not necessarily improve variational approximations.

The central challenge, therefore, is not simply constructing increasingly expressive normalising flows, but developing a stable and principled optimisation framework capable of combining heterogeneous experts while preserving computational efficiency. Surprisingly, despite extensive research on normalising flows over the past decade, relatively little attention has been devoted to the optimisation strategy governing heterogeneous flow mixtures. Existing work has predominantly focused on architectural innovation rather than on how complementary flow models should cooperate during variational optimisation.

In this work we address this problem by proposing \textbf{AMF--VI--sEMA}, an adaptive mixture framework that explicitly separates expert learning from mixture optimisation. Rather than jointly optimising heterogeneous flow parameters and mixture coefficients, each expert is first trained independently to specialise on the target distribution. Mixture weights are subsequently estimated through a lightweight likelihood-driven responsibility mechanism followed by a Simplex Exponential Moving Average (sEMA) update that operates directly on the probability simplex. This decoupled optimisation strategy stabilises training, avoids responsibility-weighted backpropagation through mixture coefficients, and enables complementary flow architectures to contribute according to their learned strengths while introducing only negligible computational overhead.

Unlike previous adaptive flow mixtures, the proposed framework is entirely architecture agnostic and requires no modification of the underlying flow models. Consequently, existing normalising flow implementations can be incorporated directly into the framework, making the approach practical for a wide range of variational inference applications.

To evaluate the proposed framework, we perform one of the most comprehensive empirical studies reported for adaptive flow mixtures. Experiments are conducted on six canonical synthetic posterior distributions together with four Bayesian inference problems exhibiting diverse posterior geometries. Evaluation extends beyond log-likelihood to include Wasserstein-$2$ distance, Maximum Mean Discrepancy (MMD), calibration analysis, optimisation stability, effective expert utilisation, computational efficiency, and robustness across repeated random initialisations. The resulting analysis provides insight not only into predictive performance but also into the optimisation behaviour of heterogeneous flow mixtures.

The principal contributions of this paper are summarised as follows.

\begin{enumerate}

\item \textbf{A decoupled optimisation framework for heterogeneous flow mixtures.}
We propose AMF--VI--sEMA, a two-stage optimisation strategy that separates expert learning from mixture adaptation through a likelihood-driven Simplex EMA weighting mechanism, substantially improving optimisation stability while remaining architecture independent.

\item \textbf{A practical adaptive weighting mechanism.}
We introduce a computationally efficient responsibility-based Simplex EMA update that estimates global mixture weights without requiring joint optimisation or per-sample gating, providing stable expert cooperation with minimal computational overhead.

\item \textbf{Comprehensive empirical evaluation.}
We evaluate the proposed framework on ten benchmark problems using complementary probabilistic, geometric, calibration, and computational metrics, demonstrating consistently robust performance across diverse posterior geometries.

\item \textbf{Analysis of optimisation dynamics.}
Beyond predictive accuracy, we analyse expert utilisation, mixture-weight evolution, optimisation stability, effective number of experts, and sensitivity to hyperparameters, providing practical insight into why heterogeneous flow mixtures succeed.

\end{enumerate}

Although specialised single-flow architectures may remain preferable for certain simple unimodal targets, the proposed framework consistently provides a more robust and reliable optimisation strategy across heterogeneous posterior landscapes without requiring prior knowledge of which flow architecture is most appropriate.

The remainder of this paper is organised as follows. Section~\ref{sec:methods} presents the proposed AMF--VI--sEMA framework. Section~\ref{sec:data_eval} introduces the benchmark datasets and evaluation protocol. Section~\ref{sec:results} reports the experimental results. Section~\ref{sec:extended} provides extended analyses of optimisation dynamics, robustness, and computational behaviour. Finally, Section~\ref{sec:conclusion} concludes the paper and discusses future research directions.

\section{Methodology}\label{sec:methods}
This section presents \emph{AMF\mbox{-}VI\mbox{-}sEMA}, a two–stage, heterogeneous mixture of normalising flows for posterior approximation. In \textbf{Stage~1}, architecturally diverse experts (e.g., \textsc{MAF}, \textsc{RealNVP}, \textsc{RBIG}) are trained \emph{independently} to specialise. In \textbf{Stage~2}, expert parameters are frozen and we learn \emph{global} mixture weights via a likelihood–driven moving–average update on the probability simplex, yielding a tractable, data–agnostic gate that reallocates mass across experts without per–sample gating. The remainder of this section formalises the mixture, details the weight–learning rule, motivates the moving–average scheme, and summarises the expert classes.

\subsection{Mixture Model Formulation}

\subsubsection{Mixture Posterior Approximation}
The variational family is a mixture of flow components
\begin{align}
    q_{\phi}(z) = \sum_{k=1}^{K} \pi_k\, q_k\!\left(z \mid \phi_k\right),\\
    \sum_{k=1}^{K}\pi_k = 1, \quad \text{where} \quad \pi_k \ge 0,
    \label{eq:mixture}
\end{align}
where $z\in\mathbb{R}^d$ denotes latent variables, $K$ is the number of components, $\pi_k$ are mixture weights, and $q_k(\cdot\mid\phi_k)$ is the $k$-th component density with parameters $\phi_k$. Each $q_k$ is instantiated as a normalising flow, i.e., an invertible transform $f_k$ that maps a simple base (typically standard Gaussian) to an expressive density via the change–of–variables formula \citep{papamakarios2021normalizing}. The mixture in \eqref{eq:mixture} accommodates multimodality and heterogeneous local geometry that single–component variational families struggle to capture.

\subsubsection{Role of Mixing Weights and Component Parameters}
The mixture weights $\pi_k$ govern the allocation of probability mass across components, modulating how the approximation covers distinct regions (e.g., secondary modes, tails) \citep{bishop2006pattern}. The component parameters $\phi_k$ shape each expert’s inductive bias (autoregressive, coupling, or Gaussianisation). In AMF\mbox{-}VI\mbox{-}sEMA, we \emph{decouple} these roles: \emph{Stage~1} fits $\{\phi_k\}$ independently for specialisation; \emph{Stage~2} adapts \emph{global} $\pi$ on the simplex using a likelihood–driven moving average. This separation avoids the instability of joint responsibility–weighted optimisation while still allowing the mixture to reallocate capacity toward components that score higher likelihood on held\mbox{-}out validation data \citep{tomczak2018vae}.

\subsection{Learning the Mixing Weights}
Classical mixtures learn component weights jointly with component parameters, but with heterogeneous flows (autoregressive, coupling, non\mbox{-}parametric Gaussianisation), joint responsibility weighted optimisation can be brittle: learning rates, curvature, and convergence scales differ markedly across experts. We therefore \emph{decouple} specialisation from weighting.

\textbf{Stage~1} trains each expert independently to specialise; \textbf{Stage~2} freezes expert parameters and learns a \emph{global} weight vector on the probability simplex via a stable, likelihood\mbox{-}driven moving average. This separation avoids coordination pathologies while still reallocating mass toward better\mbox{-}performing experts.

\subsubsection{Sequential Training Architecture}
Let $\pi=(\pi_1,\dots,\pi_K)\in\Delta^{K-1}$ with $\sum_k\pi_k=1$ and $\pi_k\ge0$. Rather than parameterising $\pi$ via unconstrained logits and backpropagating through a softmax gate, we \emph{directly} update $\pi$ on the simplex using performance\mbox{-}weighted moving averages. Two design choices ensure stability: (i) \emph{temporal separation}, where we freeze expert parameters before updating $\pi$; and (ii) \emph{simplex updates}, where we apply a convex EMA on $\pi$ with light smoothing and a small floor, preserving normalisation and preventing collapse.

\subsubsection{Two\mbox{-}Stage Training Dynamics}
\paragraph{Stage 1: Independent flow specialisation.}
Each expert $q_k(\cdot\mid\phi_k)$ is trained independently to maximise its own log\mbox{-}likelihood on data (or draws from the target simulator):
\begin{align}
  \phi_k^{\ast} \;\in\; \arg\max_{\phi_k}\; \mathbb{E}_{z\sim p_{\text{train}}}\!\left[\log q_k(z\mid\phi_k)\right].
  \label{eq:stage1}
\end{align}
Parametric flows (MAF, RealNVP) use standard gradient\mbox{-}based optimisation; RBIG is fitted by its non\mbox{-}parametric procedure. This yields complementary experts with diverse inductive biases.

\paragraph{Stage 2: Simplex EMA (sEMA) global weighting.}
With $\{\phi_k^{\ast}\}$ frozen, we update a \emph{global} mixture $\pi$ by averaging \emph{responsibilities} over held\mbox{-}out validation data and then applying an exponential moving average on the simplex. For a batch $\{z_n\}_{n=1}^{B}$ drawn from $p_{\text{val}}$,
\begin{align}
  r_{k}(z_n)
  \;=\;
  \mathrm{softmax}_{k}\!\Big(\tfrac{\log q_k(z_n\mid\phi_k^{\ast}) + \log \pi_k}{\tau}\Big),
  \\
  \bar r_k \;=\; \tfrac{1}{B}\sum_{n=1}^{B} r_k(z_n),
  \label{eq:resp}
\end{align}
where $\tau>0$ is a temperature (larger $\tau$ flattens responsibilities and reduces variance). We optionally smooth toward uniform to damp transients:
\begin{align}
  \bar r \;\leftarrow\; (1-\beta)\,\bar r \;+\; \beta\,\tfrac{1}{K}\mathbf{1},
  \qquad \beta \in [0,1].
  \label{eq:smooth}
\end{align}
We then perform a simplex EMA (with momentum $\alpha\in[0,1)$), apply a small floor to avoid zeroing, and renormalise:
\begin{align}
  \pi& \;\leftarrow\; \alpha\,\pi \;+\; (1-\alpha)\,\bar r,
  \\
  \pi& \;\leftarrow\; \max(\pi,\varepsilon),\\ \pi& \;\leftarrow\; \tfrac{\pi}{\|\pi\|_{1}},
  \label{eq:ema}
\end{align}
with $\varepsilon>0$ a small floor. We monitor the \emph{effective number of experts} $N_{\mathrm{eff}}=\exp(H(\pi))$ (where $H$ is entropy) to diagnose collapse. In practice, we use a boosting held\mbox{-}out validation set and average the resulting $\bar r$ before the EMA step, improving stability without backpropagating through $\pi$ or requiring per\mbox{-}sample gating. The update in \eqref{eq:resp}--\eqref{eq:ema} matches the implementation used in our experiments (AMF\mbox{-}VI\mbox{-}sEMA).

\subsection{Advantages of Simplex--EMA (sEMA) Global Weighting}
The proposed likelihood–driven Simplex–EMA (sEMA) update confers practical and methodological benefits over joint, responsibility–weighted training and gradient–based gating:

\begin{enumerate}[topsep=0pt,itemsep=0.2ex,partopsep=0.6ex,parsep=0.6ex]
  \item \textbf{Computational efficiency.} sEMA avoids backpropagation through the gate and any bi\mbox{-}level optimisation; each update costs $\mathcal{O}(KB)$ for $K$ experts and batch size $B$, with negligible memory overhead compared to joint training.

  \item \textbf{Numerical stability and variance control.} The EMA on the probability simplex damps oscillations, while the temperature $\tau>0$ in the softmax responsibilities (Eq.~\eqref{eq:resp}) controls concentration/variance of the estimates. Smoothing toward uniform (Eq.~\eqref{eq:smooth}) and a small floor in Eq.~\eqref{eq:ema} prevent brittle zeroing and reduce sensitivity to transient likelihood spikes.

  \item \textbf{Separation of concerns.} Decoupling specialisation (Stage~1) from global reweighting (Stage~2) avoids the coordination pathologies of joint responsibility\mbox{-}weighted optimisation with heterogeneous experts (autoregressive, coupling, non\mbox{-}parametric), where learning rates and curvature differ markedly.

  \item \textbf{Interpretability and diagnostics.} The learned $\pi$ directly reflects comparative expert performance on boosting held\mbox{-}out validation data; entropy\mbox{-}based $N_{\mathrm{eff}}=\exp(H(\pi))$ offers a simple collapse indicator and enables dataset\mbox{-}wise health checks of the mixture.

  \item \textbf{Validation\mbox{-}set assessment.} Responsibilities are averaged over $M$ mini\mbox{-}batches drawn independently and uniformly at random from a held\mbox{-}out validation split, which is kept strictly separate from the Stage~1 training data. This ensures that the gate evaluates expert likelihoods on unseen data, mitigating overfitting of $\pi$ to training idiosyncrasies and improving robustness of the learned mixture weights.

  \item \textbf{Architecture agnosticism.} sEMA treats experts as black boxes requiring only $\log q_k(z)$; it accommodates heterogeneous flows (e.g., \textsc{MAF}, \textsc{RealNVP}, \textsc{RBIG}, \textsc{NICE}, \textsc{ResFlow}) without architectural changes or shared training tricks.

  \item \textbf{Lightweight analysis.} The update $\pi \!\leftarrow\! \alpha\pi+(1{-}\alpha)\bar r$ is a contraction toward the fixed point of $\bar r$ (for fixed responsibilities), with bounded update variance under bounded log\mbox{-}likelihood noise; this supports stable trajectories and explains empirically smooth weight evolution.

  \item \textbf{Practical trade\mbox{-}offs.} sEMA does not target the same objective as joint EM or end\mbox{-}to\mbox{-}end gating; in return, it yields markedly simpler optimisation, competitive accuracy, and clearer diagnostics—well suited when reliability and ease of deployment are priorities.
\end{enumerate}

In sum, sEMA provides a stable, efficient, and interpretable mechanism for \emph{global} capacity allocation in heterogeneous mixtures, with collapse avoidance and simple diagnostics. Its effectiveness relies on expert diversity and boosting held\mbox{-}out validation data evaluation, which together enable geometry\mbox{-}aware specialisation without the fragility of joint responsibility\mbox{-}weighted training.

\subsection{Heterogeneous Flow Experts}
AMF\mbox{-}VI\mbox{-}sEMA combines architecturally distinct normalising flows whose complementary inductive biases encourage specialisation. In all cases we require only tractable $\log q_k(z)$ and sampling from each expert, which keeps Stage~2 agnostic to architectural details.

\subsubsection{Masked Autoregressive Flow (MAF)}
MAF \citep{papamakarios2017masked} models conditional dependencies via autoregressive transforms. Let $x\!\in\!\mathbb{R}^D$ and a permutation $\pi$ define the ordering. A MADE network produces $(\mu_i,\sigma_i)$ conditioned on previous dimensions, yielding
\begin{align}
z_i &= \frac{x_{\pi(i)} - \mu_i(x_{\pi(<i)})}{\sigma_i(x_{\pi(<i)})},
\\
\log\!\left|\det \tfrac{\partial z}{\partial x}\right| &= -\sum_{i=1}^{D}\log \sigma_i(\cdot).
\end{align}
Stacking layers with interleaved permutations increases expressivity. MAF tends to model sharp, directed dependencies well (useful for non\mbox{-}symmetric shapes).

\subsubsection{Real Non\mbox{-}Volume Preserving (RealNVP)}
RealNVP \citep{dinh2016realnvp} uses affine coupling with a binary mask $m$ that partitions dimensions:
\begin{align}
x_{\!A} &= z_{\!A}, \\
x_{\!B} &= z_{\!B}\odot \exp\!\big(s(z_{\!A})\big) + t(z_{\!A}), \\
\log\!\left|\det \tfrac{\partial x}{\partial z}\right| &= \sum s(z_{\!A}),
\end{align}
where $s(\cdot),t(\cdot)$ are shallow nets. Mask/permutation schedules and multi\mbox{-}scale stacks target piecewise\mbox{-}affine structure efficiently, often excelling on moderately warped densities.

\subsubsection{Rotation\mbox{-}Based Iterative Gaussianisation (RBIG)}
RBIG \citep{laparra2011iterative} alternates marginal Gaussianisation $G^{(l)}$ with orthogonal rotations $R^{(l)}$:
\begin{align}
z^{(l+1)} &= R^{(l)}\!\big(G^{(l)}(z^{(l)})\big),
\\
\log\!\left|\det \tfrac{\partial z^{(l+1)}}{\partial z^{(l)}}\right|
&= \sum_{j=1}^{D}\log g_j^{\prime\,(l)}(z^{(l)}_j),
\end{align}
where each $g_j^{(l)}$ is a univariate monotone transform (empirical CDF/\mbox{-}inverse). RBIG is non\mbox{-}parametric at the marginals and robust to heavy tails/outliers; rotations spread dependence structure across dimensions.

\subsubsection{Complementarity and specialisation}
MAF’s autoregressive conditioning captures directed, non\mbox{-}symmetric structure; RealNVP’s coupling layers yield efficient, piecewise\mbox{-}affine warps; RBIG contributes non\mbox{-}parametric marginal flexibility and rotation\mbox{-}driven mixing. Training experts independently (Stage~1) lets each converge under its own optimisation dynamics, avoiding brittle joint responsibility weighted updates across heterogeneous curvatures. The subsequent sEMA update (Stage~2) then \emph{globally} allocates probability mass according to empirical likelihood on held\mbox{-}out validation data, producing a mixture that leverages complementary priors without architectural entanglement.

\subsubsection{Additional baselines}
For the purposes of the extended experiments to evaluate the proposed AMF-VI-sEMA, we also include \textsc{NICE} (additive coupling, zero scaling), \textsc{ResFlow} (residual flows with near\mbox{-}identity transforms), and an EM\mbox{-}style mixture training procedure as references. These serve as comparators rather than experts in AMF\mbox{-}VI\mbox{-}sEMA; the Stage~2 mechanism remains unchanged and only requires access to $\log q_k$.

\subsection{Adaptive Mixture Flow Variational Inference (AMF\mbox{-}VI\mbox{-}sEMA)}
\label{sec:amfvi}

AMF\mbox{-}VI\mbox{-}sEMA is a two–stage, \emph{global} mixture of heterogeneous normalising flows.
In \textbf{Stage~1}, experts $q_k(\cdot\mid\phi_k)$ (\textsc{MAF}, \textsc{RealNVP}, \textsc{RBIG})
are trained \emph{independently} to specialise. In \textbf{Stage~2}, expert parameters are frozen and
a weight vector $\pi\in\Delta^{K-1}$ is learned by averaging temperature\mbox{-}scaled responsibilities on
held\mbox{-}out validation data and applying an exponential moving average on the simplex. No per\mbox{-}sample gating or
backpropagation through $\pi$ is used.

% ======================= TIKZ PIPELINE (COMPACT, TWO ROWS) =======================
\begin{figure}[t]
\centering
\resizebox{\linewidth}{!}{%
\begin{tikzpicture}[
  >=Latex,
  every node/.style={font=\footnotesize},
  box/.style={draw, rounded corners, thick, fill=black!3, align=center,
              minimum width=22mm, minimum height=8mm},
  flow/.style={draw, rounded corners, thick, fill=green!6, align=center,
               minimum width=20mm, minimum height=8mm},
  gate/.style={draw, rounded corners, thick, fill=red!12, align=center,
               minimum width=60mm, minimum height=18mm},
  line/.style={-Latex, very thick},
  note/.style={font=\scriptsize, inner sep=1pt}
]

% ---------- Stage 1 (top row) ----------
\node[box]  (data)   at (2.7, 0)     {Train / simulator \\ $z\!\sim\!p_{\text{train}}$};
\node[flow] (maf)    at (5.9, 1.25)  {MAF \\ $\phi_1$};
\node[flow] (rnv)    at (5.9, 0.00)  {RealNVP \\ $\phi_2$};
\node[flow] (rbig)   at (5.9,-1.25)  {RBIG \\ $\phi_3$};
\node[box]  (freeze) at (9.5, 0.00)  {Freeze experts \\ $\phi_k^{\ast}$};
\draw[line] (data) -- (maf);
\draw[line] (data) -- (rnv);
\draw[line] (data) -- (rbig);
\draw[line] (maf)  -- (freeze);
\draw[line] (rnv)  -- (freeze);
\draw[line] (rbig) -- (freeze);

% ---------- Stage 2 (bottom row) ----------
\node[box]  (val)  at (0,-3.6)    {Validation Set\\ $z\!\sim\!p_{\text{val}}$};
\node[gate] (sEMA) at (5.2,-3.6)  {Stage 2:\; Global weighting (sEMA)\\[1mm]
$\displaystyle r_k(z)=\mathrm{softmax}_k\!\Big(\tfrac{\log q_k(z\mid\phi_k^{\ast})+\log \pi_k}{\tau}\Big)$\\
$\displaystyle \bar r=\mathbb{E}_{z}[\,r(z)\,]\;\to\; (1{-}\beta)\bar r+\beta \tfrac{1}{K}\mathbf{1}$\\
$\displaystyle \pi \leftarrow \alpha\,\pi + (1{-}\alpha)\,\bar r,\;\; \pi\!\leftarrow\!\max(\pi,\varepsilon),\;\; \pi\!\leftarrow\!\pi/\|\pi\|_1$};
\node[box]  (mix)  at (12.2,-3.6) {Mixture:\\ $q(z)=\sum_k \pi_k\,q_k(z\mid\phi_k^{\ast})$\\ $N_{\text{eff}}=\exp(H(\pi))$};
\draw[line] (freeze.south) .. controls +(0,-0.9) and +(1.2,0.9) .. (sEMA.east);
\draw[line] (val)  -- (sEMA);
\draw[line] (sEMA) -- (mix);

% ---------- Labels ----------
\node[note, anchor=west] at (0, 2.1)
  {\bf Stage 1:\; Expert specialisation (independent training)};
\node[note, anchor=west] at (0,-2.0)
  {\bf Stage 2:\; Likelihood--driven global reweighting};

\end{tikzpicture}%
}
\caption{AMF\mbox{-}VI\mbox{-}sEMA pipeline. Top: independent expert training and freezing of $\phi_k^{\ast}$. Bottom: U-turn to global weight updates on held\mbox{-}out validation data via responsibilities and a simplex EMA, producing the final mixture and diagnostics ($N_{\mathrm{eff}}$).}
\label{fig:amfvi_pipeline_compact}
\end{figure}
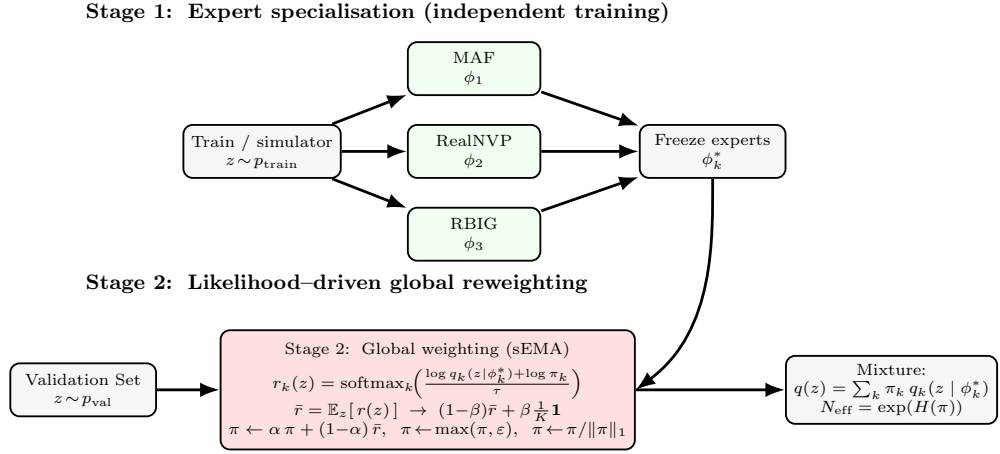

% ======================= ALGORITHM (MATCHES sEMA IMPLEMENTATION) =======================
\begin{algorithm}[t]
\caption{AMF\mbox{-}VI\mbox{-}sEMA: two-stage training with global simplex EMA}
\label{alg:amfvi}
\begin{algorithmic}[1]
\Require Experts $\{q_k(z\mid\phi_k)\}_{k=1}^{K}$, training data $p_{\text{train}}$, held\mbox{-}out validation data $p_{\text{val}}$;
temperature $\tau{>}0$, EMA momentum $\alpha\!\in[0,1)$, smoothing $\beta\!\in[0,1]$, floor $\varepsilon{>}0$, epochs $T$, batches/epoch $M$
\Statex \textbf{Stage 1: expert specialisation (independent)}
\For{$k=1$ to $K$}
  \While{not converged}
    \State sample $z\!\sim\!p_{\text{train}}$;\; update $\phi_k \leftarrow \phi_k + \eta \nabla_{\phi_k}\log q_k(z\mid\phi_k)$ \Comment{RBIG uses its own fitting}
  \EndWhile
  \State freeze $\phi_k^{\ast}\!\leftarrow\!\phi_k$
\EndFor
\State initialise $\pi^{(0)} \!\leftarrow\! \tfrac{1}{K}\mathbf{1}$
\Statex
\Statex \textbf{Stage 2: global weighting (sEMA on the simplex)}
\For{$t=0$ to $T{-}1$}
  \State $\bar r \leftarrow 0$
  \For{$m=1$ to $M$} \Comment{average responsibilities over held-out validation data}
     \State sample $\{z_n\}_{n=1}^{B}\!\sim\!p_{\text{val}}$
     \State compute $r_k(z_n)=\mathrm{softmax}_k\!\big((\log q_k(z_n\!\mid\!\phi_k^{\ast})+\log \pi_k^{(t)})/\tau\big)$
     \State $\bar r \leftarrow \bar r + \frac{1}{MB}\sum_{n=1}^{B} r(z_n)$
  \EndFor
  \State \textbf{smooth:}\; $\bar r \leftarrow (1{-}\beta)\,\bar r + \beta \tfrac{1}{K}\mathbf{1}$
  \State \textbf{EMA:}\; $\pi^{(t+\!1)} \leftarrow \alpha\,\pi^{(t)} + (1{-}\alpha)\,\bar r$
  \State \textbf{floor + renorm:}\; $\pi^{(t+\!1)}\!\leftarrow\!\max(\pi^{(t+\!1)},\varepsilon)$,\; $\pi^{(t+\!1)}\!\leftarrow\!\pi^{(t+\!1)}/\|\pi^{(t+\!1)}\|_1$
\EndFor
\State \Return mixture $q(z)=\sum_{k=1}^{K}\pi_k^{(T)}\,q_k(z\mid\phi_k^{\ast})$, and diagnostics $N_{\mathrm{eff}}=\exp(H(\pi^{(T)}))$
\end{algorithmic}
\end{algorithm}

\section{Datasets and Evaluation Protocol}
\label{sec:data_eval}

We evaluate AMF-VI-sEMA on (i) six canonical 2D synthetic posteriors spanning smooth, non-convex, and multimodal geometries, and (ii) four \emph{real} Bayesian targets (logistic, Poisson, Weibull, Real-GMM2) for which we report the same density-view metrics as in the synthetic case.

\subsection{Synthetic 2D posterior families}
We instantiate six generators with fixed parameters shared across all methods:
\begin{enumerate}[label=(\roman*)]
    \item \textit{Banana} (warped Gaussian), 
    \item \textit{X\mbox{-}shape} (crossing anisotropic modes),
    \item \textit{Bimodal} (two well\mbox{-}separated modes), 
    \item \textit{Multimodal} (five\mbox{-}component mixture),
    \item \textit{Two\mbox{-}moons} (nonlinear manifold support), and 
    \item \textit{Rings} (concentric annuli).
\end{enumerate}

Each family is instantiated once with fixed parameters (shared across methods and seeds).
Unless otherwise stated, we train on $N_{\text{train}}=300{,}000$ i.i.d.\ samples drawn
from the target distribution and evaluate on $N_{\text{eval}}=5{,}000$ held-out
i.i.d.\ samples per bootstrap iteration, drawn from a held-out test pool of
$100{,}000$ samples from the same generator.

\subsection{Real Bayesian targets (posterior in low dimension)}
We include three classical Bayesian models. To enable \emph{exact, apples\mbox{-}to\mbox{-}apples} density evaluation in 2D (and to compare with the synthetic benchmarks), we follow a standard practice:
either (a) select a two\mbox{-}parameter subvector of interest (sparse setting), or (b) learn the full posterior in $d$ and visualise/evaluate on a fixed 2D subspace (e.g., two coefficients) while reporting posterior\mbox{-}predictive metrics on all $d$. In all cases we standardise inputs and use weakly informative priors. 

By default, we fit the full posterior in dimension \(d\) and report a \emph{2D density view} on a fixed coefficient pair \((\beta_i,\beta_j)\) for visual and metric comparability with the synthetic benchmarks. In all cases we standardise inputs and use weakly informative Gaussian priors. Unless otherwise stated, all reported metrics for real targets are the density-view metrics (NLL, \(\mathrm{KL}(p\|q)\), \(W_2\), MMD).

Details of these three data sets are:
\begin{itemize}
    \item \textbf{Bayesian Logistic Regression (BLR):}
Binary outcomes \(y_n\in\{0,1\}\) with covariates \(x_n\in\mathbb{R}^{d}\),
\(p(y_n{=}1\mid x_n,\beta)=\sigma(x_n^\top\beta)\), prior \(\beta\sim\mathcal{N}(0,\lambda^{-1}I)\) with \(\lambda{=}1\).
We evaluate the posterior on a fixed 2D coefficient subspace for density-view metrics.
    \item \textbf{Bayesian Poisson Regression (BPR):} Counts $y_n\!\in\!\mathbb{N}$ with log\mbox{-}linear link,
\(
y_n \sim \mathrm{Poisson}(\lambda_n),\ \ \log \lambda_n = x_n^\top \beta,
\)
prior $\beta\sim\mathcal{N}(0,\lambda^{-1}I)$ with $\lambda{=}1$.
    \item \textbf{Weibull time\mbox{-}to\mbox{-}event (no censoring):}
Durations $t_n{>}0$ with Weibull likelihood
\(
t_n \sim \mathrm{Weibull}(k,\,\lambda_n),\ \ \log \lambda_n = x_n^\top \beta,
\)
shape $k{>}0$ (learned), rate $\lambda_n{>}0$; priors $\beta\sim\mathcal{N}(0,I)$ and
$\log k\sim\mathcal{N}(0,1)$. With censoring, we use the standard right\mbox{-}censoring likelihood; in our main experiments, we use uncensored splits.
    \item \textbf{Real Gaussian Mixture Model (Real-GMM2):}
A 2-component Gaussian mixture posterior over means $(\mu_1,\mu_2)$ fitted to real continuous data,
\(
y_i \sim \tfrac{1}{2}\mathcal{N}(\mu_1,\sigma^2) + \tfrac{1}{2}\mathcal{N}(\mu_2,\sigma^2),
\)
with priors $\mu_1,\mu_2\sim\mathcal{N}(0,\tau^2)$. The posterior is inherently multimodal due to label\mbox{-}switching symmetry ($\mu_1\leftrightarrow\mu_2$); ground-truth samples are obtained via Gibbs sampling.
\end{itemize}

\begin{table}[h]
  \centering
  \caption{Summary of datasets and evaluation views. Synthetic families and Real-GMM2 are native 2D. Regression-based real targets (BLR, BPR, Weibull) are fitted in \(d\) dimensions and evaluated on a fixed 2D coefficient subspace using the same density-view metrics.}
  \label{tab:data_summary}
  \setlength{\tabcolsep}{4pt}
  \begin{tabular}{@{}p{2.4cm}cccl@{}}
    \toprule
    \textbf{Dataset} & \textbf{Task} & \textbf{Dim ($d$)} & \textbf{Eval view} & \textbf{Metrics}\\
    \midrule
    Banana, X-Shaped, Bimodal, Multimodal, Two-moons, Rings & Density & 2 & native (2D) &
    NLL, $\mathrm{KL}(p\|q)$, $W_2$, MMD \\
    \midrule
    BLR & Binary & varies & 2D coeff.\ subspace & NLL, $\mathrm{KL}(p\|q)$, $W_2$, MMD \\
    BPR & Counts & varies & 2D coeff.\ subspace & NLL, $\mathrm{KL}(p\|q)$, $W_2$, MMD \\
    Weibull & Time-to-event & varies & 2D coeff.\ subspace & NLL, $\mathrm{KL}(p\|q)$, $W_2$, MMD \\
    Real-GMM2 & Mixture model & 2 & native (2D) & NLL, $\mathrm{KL}(p\|q)$, $W_2$, MMD \\
    \bottomrule
  \end{tabular}
\end{table}

\subsection{Baselines, training, and common settings}
All flow baselines use their \emph{standard} objectives and architectures with a unified budget across datasets to isolate shape effects. We include AMF\mbox{-}VI\mbox{-}sEMA expert models of \textsc{MAF}, \textsc{RealNVP}, \textsc{RBIG} along with \textsc{NICE}, \textsc{ResFlow}, and an EM\mbox{-}mixing reference.
For AMF\mbox{-}VI\mbox{-}sEMA:
\begin{itemize}
    \item \textbf{Stage~1} trains experts independently (same epochs/lr/batch across datasets);
    \item \textbf{Stage~2} uses mini-batches drawn from the held-out validation split with defaults
        \[(\tau,\alpha,\beta,M)=(1.1,\;0.9,\;10^{-5},\;2)\]
        where $M{=}2$ validation mini-batches are averaged per responsibility update (Eq.~23).
    \item We log weight trajectories and the effective number of experts $N_{\text{eff}}{=}\exp(H(\pi))$.
\end{itemize}

We report complementary metrics in Table~\ref{tab:metric_defs} for NLL, $\mathrm{KL}(p\|q)$, $W_2$, and MMD (unbiased and biased).
All quantities are estimated from i.i.d.\ samples:
\emph{NLL} via $\tfrac{1}{N}\sum_{i}\!-\log q(\tilde z_i)$ with $\{\tilde z_i\}\!\sim\!p$;
\emph{$\mathrm{KL}(p\|q)$} via KDE\mbox{-}based quadrature on a fixed grid (Scott's rule; $\varepsilon$ floor);
\emph{$W_2$} as the empirical OT distance between $N$ samples (Sinkhorn $\varepsilon{=}10^{-2}$; debiased);
\emph{MMD} with Gaussian kernels and the median heuristic, reporting both the
unbiased U\mbox{-}statistic (MMD-u) and biased V\mbox{-}statistic (MMD-b).
All numbers are averaged over $R{=}10$ bootstrap iterations; each iteration independently resamples $N_{\text{eval}}{=}5{,}000$ samples from a target pool of $25{,}000$ to reduce evaluation variance. We report mean values for scalar metrics. Hyperparameters (epochs, batch size, learning rate) are kept constant across datasets unless a baseline's reference implementation prescribes otherwise.

\begin{table}[h]
  \centering
  \caption{Metrics, expressions, and rationale. $^*$ ``u'' = unbiased U\mbox{-}statistic; ``b'' = biased V\mbox{-}statistic (lower variance).}
  \label{tab:metric_defs}
  \setlength{\tabcolsep}{3pt}
  % \scriptsize
  \begin{tabular}{@{}p{2.3cm}l p{6.5cm}@{}}
    \toprule
    \textbf{Metric} & \textbf{Expression} & \textbf{Why use here} \\
    \midrule
    NLL & $\displaystyle -\,\mathbb{E}_{z\sim p}\big[\log q(z)\big] \;=\; H(p)+\mathrm{KL}(p\|q)$ 
        & Proxy for the VI objective; penalises mass loss/mode dropping. \\
    \addlinespace[2pt]
    $\mathrm{KL}(p\|q)$ & $\displaystyle \mathbb{E}_{z\sim p}\!\left[\log\tfrac{p(z)}{q(z)}\right]$
        & Mass\mbox{-}covering divergence aligned with ELBO optimisation. \\
    \addlinespace[2pt]
    $W_2$ & $\displaystyle \Big(\inf_{\gamma\in\Pi(p,q)} \mathbb{E}_{(x,y)\sim\gamma}\|x-y\|_2^2\Big)^{\!1/2}$
        & Transport/geometry aware; captures displacement and shape. \\
    \addlinespace[2pt]
    MMD (u/b)$^*$ & $\displaystyle \mathrm{MMD}^2_k(p,q)=\mathbb{E}_{p,p}k+\mathbb{E}_{q,q}k-2\,\mathbb{E}_{p,q}k$
        & Nonparametric two\mbox{-}sample discrepancy complementary to NLL/KL/OT. \\
    \bottomrule
  \end{tabular}
\end{table}

\section{Results and Analysis}\label{sec:results}
We evaluate \emph{AMF\mbox{-}VI\mbox{-}sEMA} against single\mbox{-}flow baselines across six canonical 2D posterior families spanning unimodal, curved, and strongly multimodal geometries, providing a stringent test of robustness to distributional shape. Performance is assessed using five complementary metrics (Sec.~\ref{sec:data_eval}), with complete numerical results reported in Table~\ref{tab:results}. Qualitative density visualisations are shown in Fig.~\ref{fig:contours}, while the learned global weights and corresponding effective number of experts are presented in Fig.~\ref{fig:weights}.

For clarity, we organise the quantitative discussion into three parts: (i) \emph{overall likelihood}, assessed via negative log\mbox{-}likelihood (NLL); (ii) \emph{transport behaviour}, measured by the Wasserstein\mbox{--}2 distance ($W_2$); and (iii) \emph{divergence and discrepancy measures}, including $\mathrm{KL}(p\|q)$ and MMD (both unbiased and biased variants), which together provide complementary perspectives on approximation quality across datasets. Unless otherwise stated, all results are reported as means over repeated runs.

\subsection{Overall likelihood}
Across the six synthetic families and four real\;/\;low\mbox{-}dimensional posteriors, \textbf{AMF\mbox{-}VI\mbox{-}sEMA} delivers consistently strong NLL performance. On synthetics it attains the best or near\mbox{-}best NLL in every case: best on \emph{Rings} (2.218, narrowly ahead of \textsc{RealNVP} at 2.220), tied best on \emph{X\mbox{-}shaped} (3.134, matching \textsc{RealNVP}), and second on \emph{Banana} (3.212, behind \textsc{RealNVP} at 3.198), \emph{Bimodal} (2.947, behind \textsc{RealNVP} at 2.946), and \emph{Two\mbox{-}moons} (1.012, behind \textsc{RealNVP} at 1.008 but matching \textsc{MAF}). On \emph{Multimodal}, \textsc{RealNVP} posts the lowest NLL (3.231), but \emph{sEMA} (3.238) and \emph{AMF\mbox{-}VI} (3.240) remain closely competitive.

Importantly, \emph{sEMA} consistently improves NLL over the original \emph{AMF\mbox{-}VI} across all six synthetics (e.g., \emph{Rings} 2.218 vs.\ 2.350; \emph{Two\mbox{-}moons} 1.012 vs.\ 1.074; \emph{X\mbox{-}shaped} 3.134 vs.\ 3.172). 

On the \emph{real} targets, \emph{sEMA} remains competitive across all four datasets: near\mbox{-}best on BLR ($-3.480$, within 0.002 of \textsc{MAF}'s $-3.482$) and BPR ($-4.487$, within 0.005 of \textsc{MAF}'s $-4.492$), second on Weibull ($-6.456$, behind \textsc{NICE} at $-6.490$), and second on Real\mbox{-}GMM2 ($-1.619$, behind \textsc{MAF} at $-1.636$). In all cases \emph{sEMA} outperforms the base \emph{AMF\mbox{-}VI}, indicating that the sEMA weight\mbox{-}update mechanism yields consistent improvements without degradation.

\subsection{Transport behaviour and discrepancies}
While likelihood dominates VI objectives, geometry\mbox{-}aware and two\mbox{-}sample metrics reveal important robustness differences. On synthetics, \emph{sEMA} consistently attains top\mbox{-}tier $W_2$ and MMD across all families. On \emph{X\mbox{-}shaped}, it matches \textsc{RealNVP} for the second\mbox{-}best $W_2$ (0.157, behind \textsc{RBIG} at 0.148) with competitive MMD\mbox{-}u/b (0.007/0.019). On \emph{Two\mbox{-}moons}, it achieves competitive $W_2$ (0.055, behind \textsc{MAF} at 0.049, \textsc{RBIG} at 0.052, and \textsc{RealNVP} at 0.053) alongside low MMD (0.006/0.015). On \emph{Multimodal}, \emph{sEMA} ties \textsc{MAF} for the best $W_2$ (0.133) with competitive MMD\mbox{-}u/b (0.005/0.017).

For \emph{Rings}, \textsc{RBIG} excels in geometric fidelity ($W_2{=}0.118$), with \textsc{RealNVP} (0.130) and \textsc{MAF} (0.131) close behind; \emph{sEMA} remains competitive ($W_2{=}0.149$) while achieving the second\mbox{-}best MMD\mbox{-}u (0.004). 

Notably, methods that appear strong on NLL alone can be unstable in geometry: \textsc{NICE} posts competitive NLL on some datasets but shows substantially worse $W_2$ and MMD elsewhere (e.g., \emph{X\mbox{-}shaped} $W_2{=}0.579$, \emph{Rings} $W_2{=}0.465$, \emph{Two\mbox{-}moons} $W_2{=}0.214$). \textsc{ResFlow} is similarly brittle in transport across all synthetics, while \textsc{EM\mbox{-}Mix} fails catastrophically (e.g., \emph{Two\mbox{-}moons} $W_2{=}24.278$, \emph{Rings} $W_2{=}17.569$, \emph{Multimodal} $W_2{=}2.417$).

On the real targets, \emph{sEMA} yields uniformly small $W_2$ and MMD alongside \textsc{MAF}, \textsc{RealNVP}, and \textsc{RBIG} (e.g., BLR $W_2{=}0.005$, MMD\mbox{-}u/b $0.000/0.001$; BPR $W_2{=}0.003$; Weibull $W_2{=}0.001$; Real\mbox{-}GMM2 $W_2{=}0.013$), while \textsc{ResFlow} and \textsc{EM\mbox{-}Mix} fail completely.

\subsection{AMF\mbox{-}VI\mbox{-}sEMA vs.\ AMF\mbox{-}VI and single\mbox{-}flow baselines}
Relative to the original two\mbox{-}stage \emph{AMF\mbox{-}VI}, \emph{sEMA} improves or matches NLL across all synthetic families and real targets, while KL and MMD remain essentially unchanged — indicating that the sEMA update yields consistent likelihood gains without sacrificing distributional fidelity (e.g., \emph{Rings} 2.218 vs.\ 2.350; \emph{Two\mbox{-}moons} 1.012 vs.\ 1.074).

Against single\mbox{-}flow baselines, \emph{sEMA} avoids the pathological failures that affect specialised models. \textsc{ResFlow} degrades severely in transport across all datasets, and \textsc{EM\mbox{-}Mix} is unstable on curved geometries. Where individual models achieve isolated metric wins, they typically trade off performance elsewhere: \textsc{RBIG} is strong in $W_2$ on ring\mbox{-}like structures but weaker in NLL; \textsc{NICE} attains competitive NLL on some synthetics yet exhibits large transport errors (e.g., \emph{X\mbox{-}shaped} $W_2{=}0.579$). KL divergences further reinforce this pattern: \textsc{NICE} and \textsc{ResFlow} show substantially worse KL on several datasets, whereas \emph{sEMA} and \emph{AMF\mbox{-}VI} remain consistently well-calibrated.

\textit{Summarising Table \ref{tab:results},}
\emph{AMF\mbox{-}VI\mbox{-}sEMA} achieves consistently strong likelihood performance while maintaining robust geometry and discrepancy scores across both synthetic and real posteriors. Compared to \emph{AMF\mbox{-}VI}, it provides stable NLL improvements without regression on other metrics; compared to single\mbox{-}flow baselines, it trades isolated metric wins for reliable, across\mbox{-}metric robustness.

\begin{table}[htbp]
  \centering
  \caption{Quantitative results on ten posterior families (five synthetic, five real/low-dimensional). \textbf{Bold} values denote the best (lowest) per row; \underline{underline} denotes the second-best. Ties share the same rank. EM-Mix KL values marked $0.000$ on real targets are numerical artifacts (degenerate fits) and are not bolded.}
  \label{tab:results}
  \setlength\tabcolsep{3.5pt}
  \scriptsize
  \begin{tabular}{cl|c|c|ccccc c}
    \toprule
    Data & Metrics & AMF-VI-sEMA & AMF-VI & RealNVP & MAF & RBIG & NICE & ResFlow & EM-Mix \\
    \toprule
    %% ── Banana ──────────────────────────────────────────────────────────────
    \multirow{5}{*}{\begin{turn}{45}Banana\end{turn}}
          & NLL   & 3.212 & 3.217 & \textbf{3.198} & \underline{3.201} & 3.261 & 3.231 & 3.456 & 3.384 \\
          & KL    & \textbf{0.014} & \textbf{0.014} & \textbf{0.014} & \textbf{0.014} & \textbf{0.014} & \underline{0.027} & 0.286 & 0.831 \\
          & $W_2$ & 0.162 & 0.167 & \underline{0.150} & 0.157 & \textbf{0.141} & 0.200 & 0.796 & 2.179 \\
          & MMD-u & \textbf{0.006} & \textbf{0.006} & \underline{0.007} & 0.008 & \underline{0.007} & 0.012 & 0.178 & 0.161 \\
          & MMD-b & \textbf{0.018} & \textbf{0.018} & \underline{0.019} & 0.020 & \underline{0.019} & 0.030 & 0.180 & 0.163 \\\midrule
    %% ── X-Shaped ─────────────────────────────────────────────────────────────
    \multirow{5}{*}{\begin{turn}{45}X-Shaped\end{turn}}
          & NLL   & \textbf{3.134} & 3.172 & \textbf{3.134} & \underline{3.145} & 3.465 & 3.535 & 3.558 & 3.324 \\
          & KL    & \textbf{0.014} & \textbf{0.014} & \underline{0.015} & \underline{0.015} & \underline{0.015} & 0.295 & 0.318 & 0.657 \\
          & $W_2$ & \underline{0.157} & 0.163 & \underline{0.157} & 0.160 & \textbf{0.148} & 0.579 & 0.695 & 1.935 \\
          & MMD-u & \textbf{0.007} & \textbf{0.007} & \underline{0.008} & 0.009 & \textbf{0.007} & 0.114 & 0.176 & 0.181 \\
          & MMD-b & \textbf{0.019} & \textbf{0.019} & \textbf{0.019} & \underline{0.020} & \textbf{0.019} & 0.118 & 0.179 & 0.183 \\\midrule
    %% ── Bimodal ──────────────────────────────────────────────────────────────
    \multirow{5}{*}{\begin{turn}{45}Bimodal\end{turn}}
          & NLL   & \underline{2.947} & 2.948 & \textbf{2.946} & 2.954 & 2.963 & 3.003 & 3.262 & 3.504 \\
          & KL    & \underline{0.013} & \underline{0.013} & \textbf{0.012} & \textbf{0.012} & 0.014 & 0.049 & 0.268 & 0.844 \\
          & $W_2$ & 0.115 & 0.120 & \underline{0.112} & \textbf{0.110} & \underline{0.112} & 0.216 & 0.625 & 1.474 \\
          & MMD-u & 0.008 & 0.008 & \underline{0.007} & \textbf{0.006} & 0.008 & 0.036 & 0.188 & 0.295 \\
          & MMD-b & \textbf{0.018} & \textbf{0.018} & \textbf{0.018} & \underline{0.019} & \underline{0.019} & 0.045 & 0.190 & 0.296 \\\midrule
    %% ── Multimodal-5 ─────────────────────────────────────────────────────────
    \multirow{5}{*}{\begin{turn}{45}Multimodal\end{turn}}
          & NLL   & 3.238 & 3.240 & \textbf{3.231} & \underline{3.237} & 3.272 & 3.293 & 3.463 & 3.440 \\
          & KL    & \underline{0.015} & \underline{0.015} & \textbf{0.014} & \underline{0.015} & \underline{0.015} & 0.046 & 0.225 & 1.115 \\
          & $W_2$ & \textbf{0.133} & 0.138 & 0.135 & \textbf{0.133} & \underline{0.134} & 0.264 & 0.760 & 2.417 \\
          & MMD-u & \underline{0.005} & \underline{0.005} & \textbf{0.004} & \textbf{0.004} & \underline{0.005} & 0.036 & 0.156 & 0.233 \\
          & MMD-b & \textbf{0.017} & \textbf{0.017} & \textbf{0.017} & \textbf{0.017} & \underline{0.018} & 0.046 & 0.159 & 0.235 \\\midrule
    %% ── Two-moons ────────────────────────────────────────────────────────────
    \multirow{5}{*}{\begin{turn}{45}Two-moons\end{turn}}
          & NLL   & \underline{1.012} & 1.074 & \textbf{1.008} & \underline{1.012} & 1.419 & 1.219 & 2.679 & 1.429 \\
          & KL    & \textbf{0.004} & \textbf{0.004} & \underline{0.005} & \textbf{0.004} & \underline{0.005} & 0.053 & 1.030 & 0.782 \\
          & $W_2$ & 0.055 & 0.056 & 0.053 & \textbf{0.049} & \underline{0.052} & 0.214 & 1.076 & 24.278 \\
          & MMD-u & 0.006 & 0.006 & 0.008 & \textbf{0.003} & \underline{0.005} & 0.068 & 0.351 & 0.660 \\
          & MMD-b & 0.015 & 0.015 & 0.015 & \textbf{0.012} & \underline{0.014} & 0.071 & 0.352 & 0.660 \\\midrule
    %% ── Rings ────────────────────────────────────────────────────────────────
    \multirow{5}{*}{\begin{turn}{45}Rings\end{turn}}
          & NLL   & \textbf{2.218} & 2.350 & \underline{2.220} & 2.496 & 2.580 & 3.433 & 3.418 & 2.545 \\
          & KL    & \textbf{0.004} & \textbf{0.004} & \textbf{0.004} & \textbf{0.004} & \textbf{0.004} & \underline{0.154} & 0.167 & 2.145 \\
          & $W_2$ & 0.149 & 0.150 & \underline{0.130} & 0.131 & \textbf{0.118} & 0.465 & 0.518 & 17.569 \\
          & MMD-u & \underline{0.004} & \underline{0.004} & \underline{0.004} & \textbf{0.003} & \textbf{0.003} & 0.046 & 0.038 & 0.405 \\
          & MMD-b & 0.018 & 0.018 & \underline{0.017} & \underline{0.017} & \textbf{0.016} & 0.054 & 0.048 & 0.406 \\\midrule\midrule
    %% ── BLR ──────────────────────────────────────────────────────────────────
    \multirow{5}{*}{\begin{turn}{45}BLR\end{turn}}
          & NLL   & $\underline{-3.480}$ & $-3.414$ & $-2.063$ & $\mathbf{-3.482}$ & $-3.470$ & $-3.433$ & $2.091$ & $-0.708$ \\
          & KL    & \underline{0.016} & \underline{0.016} & \textbf{0.015} & \underline{0.016} & \underline{0.016} & 0.057 & 5.340 & $0.000$ \\
          & $W_2$ & \textbf{0.005} & \textbf{0.005} & \textbf{0.005} & \textbf{0.005} & \textbf{0.005} & \underline{0.013} & 1.559 & 54.806 \\
          & MMD-u & \textbf{0.000} & \textbf{0.000} & \textbf{0.000} & \underline{0.001} & \textbf{0.000} & 0.011 & 0.647 & 0.993 \\
          & MMD-b & \textbf{0.001} & \textbf{0.001} & \textbf{0.001} & \textbf{0.001} & \textbf{0.001} & \underline{0.012} & 0.647 & 0.993 \\\midrule
    %% ── BPR ──────────────────────────────────────────────────────────────────
    \multirow{5}{*}{\begin{turn}{45}BPR\end{turn}}
          & NLL   & $\underline{-4.487}$ & $-4.451$ & $-2.126$ & $\mathbf{-4.492}$ & $-4.477$ & $-4.470$ & $2.077$ & $-1.222$ \\
          & KL    & \textbf{0.018} & \textbf{0.018} & \underline{0.019} & \textbf{0.018} & 0.020 & 0.038 & 4.428 & $0.000$ \\
          & $W_2$ & \textbf{0.003} & \textbf{0.003} & \textbf{0.003} & \textbf{0.003} & \textbf{0.003} & \underline{0.005} & 1.559 & 42.048 \\
          & MMD-u & \textbf{0.000} & \textbf{0.000} & \textbf{0.000} & \textbf{0.000} & \textbf{0.000} & \underline{0.002} & 0.638 & 0.986 \\
          & MMD-b & \underline{0.001} & \underline{0.001} & \underline{0.001} & \underline{0.001} & \textbf{0.000} & 0.002 & 0.638 & 0.987 \\\midrule
    %% ── Weibull ──────────────────────────────────────────────────────────────
    \multirow{5}{*}{\begin{turn}{45}Weibull\end{turn}}
          & NLL   & $-6.456$ & $\underline{-6.458}$ & $-2.157$ & $-6.436$ & $-6.441$ & $\mathbf{-6.490}$ & $2.397$ & $-1.825$ \\
          & KL    & \textbf{0.016} & \textbf{0.016} & \underline{0.017} & \underline{0.017} & \textbf{0.016} & 0.035 & 16.201 & $0.000$ \\
          & $W_2$ & \textbf{0.001} & \textbf{0.001} & \textbf{0.001} & \textbf{0.001} & \textbf{0.001} & \underline{0.002} & 1.967 & 64.779 \\
          & MMD-u & \textbf{0.000} & \textbf{0.000} & \textbf{0.000} & \textbf{0.000} & \textbf{0.000} & \underline{0.001} & 0.741 & 1.000 \\
          & MMD-b & \textbf{0.000} & \textbf{0.000} & \textbf{0.000} & \textbf{0.000} & \textbf{0.000} & \underline{0.001} & 0.741 & 1.000 \\\midrule
    %% ── Real-GMM2 ────────────────────────────────────────────────────────────
    \multirow{5}{*}{\begin{turn}{45}Real-GMM2\end{turn}}
          & NLL   & $\underline{-1.619}$ & $-1.608$ & $-1.533$ & $\mathbf{-1.636}$ & $-1.595$ & $-1.614$ & $2.038$ & $-0.278$ \\
          & KL    & \textbf{0.014} & \textbf{0.014} & \underline{0.015} & \underline{0.015} & \textbf{0.014} & 0.027 & 3.584 & $0.000$ \\
          & $W_2$ & \textbf{0.013} & \underline{0.014} & \underline{0.014} & \underline{0.014} & \underline{0.014} & 0.019 & 1.351 & 53.923 \\
          & MMD-u & \underline{0.004} & \underline{0.004} & 0.007 & 0.006 & 0.006 & \textbf{0.001} & 0.586 & 0.960 \\
          & MMD-b & \underline{0.006} & \underline{0.006} & 0.009 & 0.008 & 0.008 & \textbf{0.005} & 0.586 & 0.960 \\\bottomrule
  \end{tabular}
\end{table}

\begin{figure}[ht]
    \centering
    \includegraphics[width=\linewidth]{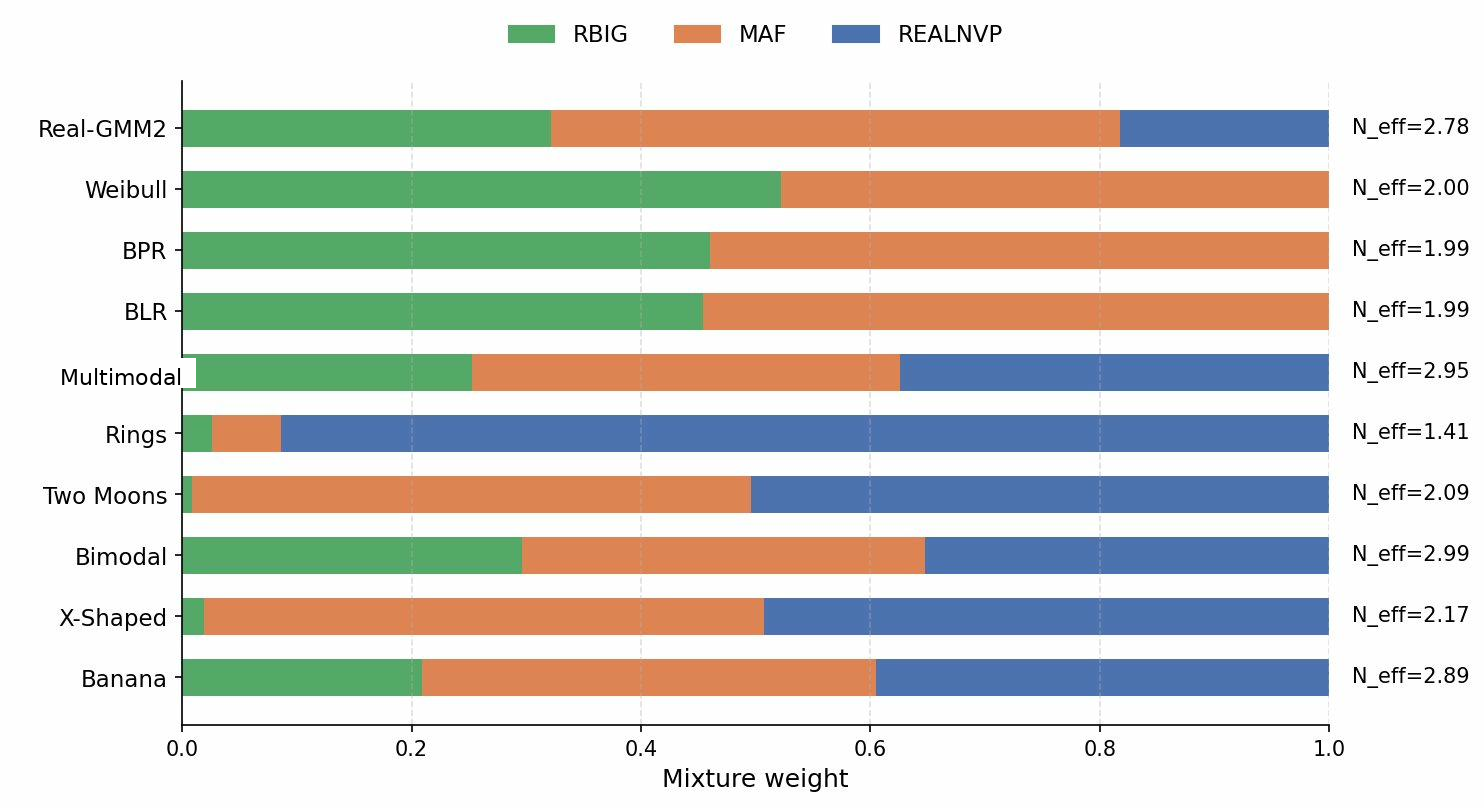}
    \caption{Learned \emph{global} mixture weights for \textbf{AMF-VI-sEMA} across six synthetic and four real datasets. Right-side labels report \(N_{\text{eff}}=\exp(H(\pi))\), indicating the effective expert count (range \([1,3]\); no collapse when \(N_{\text{eff}}\!\gg\!1\)). The gate adapts to target geometry: synthetics retain broad participation (\(N_{\text{eff}}\!\geq\!1.41\)), while real regression targets suppress \textsc{RealNVP} in favour of a \textsc{MAF}/\textsc{RBIG} split.}

\label{fig:weights}
\end{figure}

Figure~\ref{fig:weights} shows the learned \emph{global} weights for \textsc{RealNVP}, \textsc{MAF}, and \textsc{RBIG} under \textbf{AMF-VI-sEMA} across all ten datasets (six synthetics + four real). The gate adapts to geometry rather than collapsing to a single expert. On the synthetics, the allocation is broadly sensible: for \textit{Bimodal} the gate distributes mass nearly equally (\(0.352/0.351/0.297\) for \textsc{RealNVP}/\textsc{MAF}/\textsc{RBIG}); on \textit{Multimodal} the split is similarly balanced (\(0.374/0.374/0.253\)); on \textit{Two-moons} the gate almost entirely dismisses \textsc{RBIG} (\(0.009\)) in favour of a near-equal split between \textsc{RealNVP} (\(0.504\)) and \textsc{MAF} (\(0.487\)); \textit{Rings} is heavily dominated by \textsc{RealNVP} (\(0.914\)) with small residual shares for \textsc{MAF} (\(0.060\)) and \textsc{RBIG} (\(0.026\)); \textit{X-shaped} is nearly two-way between \textsc{RealNVP} (\(0.493\)) and \textsc{MAF} (\(0.488\)) with \textsc{RBIG} almost zeroed out (\(0.019\)); and \textit{Banana} is close to a uniform two-way split (\(0.395/0.397/0.209\)). 

On the \emph{real} Bayesian targets, the gate reflects the different likelihood geometries: \textit{BLR} assigns virtually no mass to \textsc{RealNVP} ($\approx 1.3\times10^{-5}$) and splits between \textsc{MAF} (\(0.545\)) and \textsc{RBIG} (\(0.455\)); \textit{BPR} follows the same pattern with negligible \textsc{RealNVP} ($\approx 1.0\times10^{-5}$) and a near-equal \textsc{MAF}/\textsc{RBIG} split (\(0.540/0.460\)); \textit{Weibull} similarly suppresses \textsc{RealNVP} ($\approx 1.0\times10^{-5}$) and divides mass between \textsc{RBIG} (\(0.522\)) and \textsc{MAF} (\(0.478\)); and \textit{Real-GMM2} partially recovers a role for \textsc{RealNVP} (\(0.183\)), with the remainder split between \textsc{MAF} (\(0.496\)) and \textsc{RBIG} (\(0.322\)). The consistent near-zero \textsc{RealNVP} weight on three of four real targets suggests the gate learns that RealNVP's coupling-layer inductive bias is poorly suited to regression-style posteriors. These patterns align with the quantitative results in Table~\ref{tab:results}, where \textbf{AMF-VI-sEMA} delivers strong likelihoods while avoiding the sharp regressions observed in some single-flow baselines.

To quantify how many experts are effectively used, we report \(N_{\text{eff}}(\pi)=\exp(H(\pi))\in[1,3]\), shown as right-side labels in Fig.~\ref{fig:weights}. Values are close to three on \textit{Bimodal} (\(2.99\)), \textit{Multimodal} (\(2.95\)), and \textit{Banana} (\(2.89\))—indicating broad participation across all three experts—while \textit{Real-GMM2} (\(2.78\)) also retains high diversity. More selective but still multi-expert behaviour is observed on \textit{X-shaped} (\(2.17\)) and \textit{Two-moons} (\(2.09\)), where two experts dominate with a near-zero \textsc{RBIG} share. \textit{Rings} shows the strongest specialisation among synthetics (\(N_{\text{eff}}{=}1.41\)), consistent with its near-total concentration on \textsc{RealNVP}. The real regression targets exhibit moderate specialisation without collapse: \textit{BLR} and \textit{BPR} both reach \(1.99\) and \textit{Weibull} \(2.00\), reflecting the two-way \textsc{MAF}/\textsc{RBIG} split with suppressed \textsc{RealNVP}. Overall, the gate exploits complementary inductive biases, adapts emphasis to the target family, and maintains diversity (\(N_{\text{eff}}\!>\!1.4\) in all cases), which helps explain the stable, across-family NLL gains observed for \textbf{AMF-VI-sEMA}.

\begin{figure*}[t!]
    \centering
    \includegraphics[width=\linewidth]{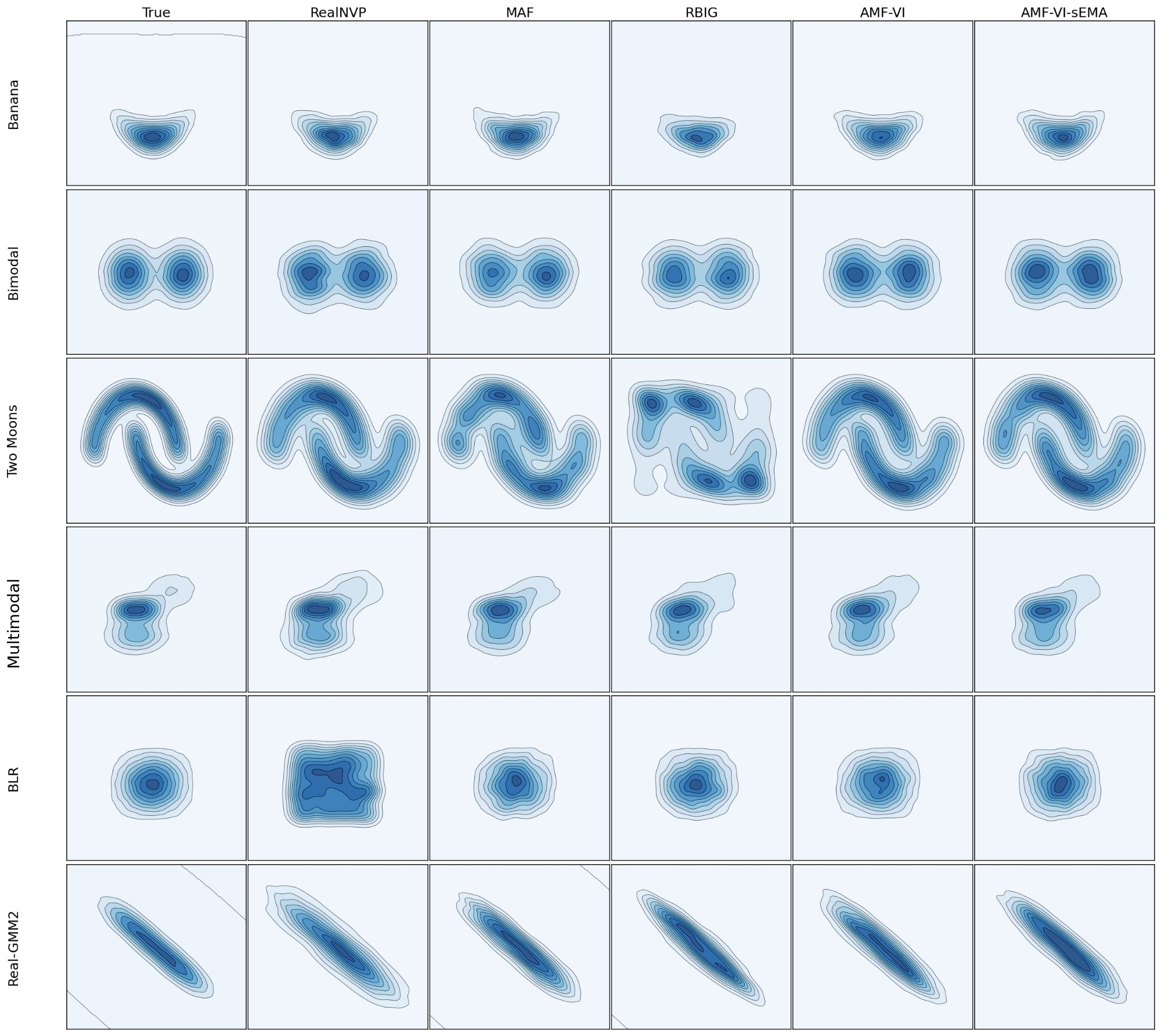}
\caption{\textbf{Qualitative density contours across six targets (top to bottom):}
\emph{Banana}, \emph{Bimodal}, \emph{Two-moons}, \emph{Multimodal}, \emph{BLR}, and \emph{Real-GMM2}.
Each row shows the ground-truth (left) and model contours for \textsc{RealNVP}, \textsc{MAF}, \textsc{RBIG}, \textbf{AMF-VI}, and \textbf{AMF-VI-sEMA} (right).
Across datasets, sEMA preserves geometric structure (curved banana shape, separated bimodal peaks, interleaved crescent manifolds, clustered multi-mode mass, compact unimodal posterior, elongated correlation ridge) with fewer bridges/leaks and less over-/under-smoothing than single flows, and typically sharpens residual artefacts observed with AMF-VI, aligning with the quantitative gains in Table~\ref{tab:results}.}
\label{fig:contours}
\end{figure*}

Figure~\ref{fig:contours} presents qualitative density contours for six representative targets—(from top to bottom) \emph{Banana}, \emph{Bimodal}, \emph{Two-moons}, \emph{Multimodal}, \emph{BLR}, and \emph{Real-GMM2}—contrasting \textit{AMF\mbox{-}VI\mbox{-}sEMA} against single-flow baselines (\textsc{RealNVP}, \textsc{MAF}, \textsc{RBIG}) and its predecessor \textit{AMF\mbox{-}VI} \citep{wiriyapong2025adaptive}; we interpret these together with Table~\ref{tab:results}.

On the \emph{Banana} distribution, all methods broadly recover the curved shape, but \textsc{RealNVP} slightly under-covers the tails and \textsc{RBIG} rounds the outer contours. \textit{sEMA} maintains a clean concave ridge with well-placed contour spacing, consistent with its competitive NLL and $W_2$.

For the \emph{Bimodal} case, the sEMA gate keeps modes clearly separated with near-equal peak density. \textsc{RealNVP} over-concentrates the inner region, \textsc{MAF} inflates central mass creating a bridge between modes, and \textsc{RBIG} spreads density between peaks. Relative to \textit{AMF\mbox{-}VI}, sEMA reduces inter-mode leakage.

On \emph{Two-moons}, \textit{sEMA} faithfully traces the interleaved crescent structure. \textsc{RealNVP} over-smooths the manifold, \textsc{MAF} loses the narrow gap between crescents, and \textsc{RBIG} fragments into disconnected blobs. sEMA closely matches \textit{AMF\mbox{-}VI} visually while posting a notably lower NLL (1.012 vs.\ 1.074).

For the \emph{Multimodal} dataset, single flows either merge adjacent modes (\textsc{RealNVP}, \textsc{MAF}) or diffuse density (\textsc{RBIG}), whereas sEMA maintains distinct, compact modes with contours closely matching the ground truth.

In the \emph{BLR} setting, where the posterior is compact and unimodal, \textsc{RealNVP} clearly over-disperses with a square-shaped spread, while \textsc{MAF}, \textsc{RBIG}, \textit{AMF\mbox{-}VI}, and \textit{sEMA} all recover the elliptical shape well, consistent with their uniformly small $W_2$ and MMD.

Finally, for \emph{Real-GMM2}, the target forms an elongated, highly correlated ridge. \textsc{MAF} and \textsc{RBIG} slightly over-extend the tails, while \textit{sEMA} and \textit{AMF\mbox{-}VI} both capture the orientation and scale faithfully, with \textit{sEMA} achieving the best $W_2$ (0.013) on this dataset.

Overall, across all six displayed targets, \textit{sEMA} adapts global weights to the geometry at hand, avoiding the under-/over-dispersion trade-offs of single flows and removing minor artefacts remaining in \emph{AMF\mbox{-}VI}. These qualitative trends mirror the quantitative improvements reported in Table~\ref{tab:results}.

\subsection{Summary of findings}
Across all experiments, a consistent pattern emerges: \emph{AMF\mbox{-}VI\mbox{-}sEMA} provides reliable performance across diverse posterior geometries, rather than excelling only in isolated cases. On synthetic datasets, it achieves best or near\mbox{-}best likelihood while maintaining strong transport and discrepancy metrics, indicating that improved density estimation does not come at the cost of geometric fidelity. On real and low\mbox{-}dimensional posteriors, where simpler unimodal structure often favours specialised models such as \textsc{MAF}, \emph{sEMA} remains consistently competitive, demonstrating that the mixture does not degrade when multimodality is less pronounced. 

Crucially, the global weighting mechanism adapts to the underlying geometry by selectively combining complementary experts, as reflected in the learned weight distributions and effective number of experts. This results in a model that avoids the failure modes of individual flows and the instability of joint mixture training, while requiring minimal additional complexity. Overall, these findings position \emph{AMF\mbox{-}VI\mbox{-}sEMA} as a robust and practical framework for posterior approximation across heterogeneous settings.

\section{Extended Analysis}
\label{sec:extended}

Beyond aggregate likelihood and discrepancy metrics, we conduct an extended analysis to characterise the internal behaviour, stability, and robustness of the proposed \emph{AMF–VI–sEMA} framework. 
This section investigates how the stochastic EMA gating mechanism influences mixture dynamics, hyperparameter sensitivity, and cross-seed reproducibility across a diverse set of posterior benchmarks. 
Our goal is to move beyond performance comparison and provide empirical evidence for the interpretability, temporal stability, and reliability of the mixture under varying conditions.

Specifically, Section~\ref{sec:51} examines expert responsibility evolution and gate volatility over training time, revealing how sEMA regularisation smooths and stabilises expert transitions. 
Section~\ref{sec:52} quantifies the effect of hyperparameters $\tau$, $\alpha$, and $M$ on effective expert utilisation and volatility–churn characteristics. 
Finally, Section~\ref{sec:53} evaluates robustness to random initialisation, measuring variance and coefficient of variation across metrics such as NLL, KL divergence, Wasserstein distance, and MMD. 
% Together, these analyses provide a holistic view of the \emph{AMF–VI–sEMA} model’s stability and generalisation behaviour beyond standard likelihood evaluation.
 
\subsection{Gate dynamics (sEMA): oscillatory, damped, stationary}\label{sec:51}

\begin{figure*}[htbp]
  \centering
  % Multimodal-5
  \begin{subfigure}[t]{0.48\linewidth}
    \centering
    \includegraphics[width=\linewidth]{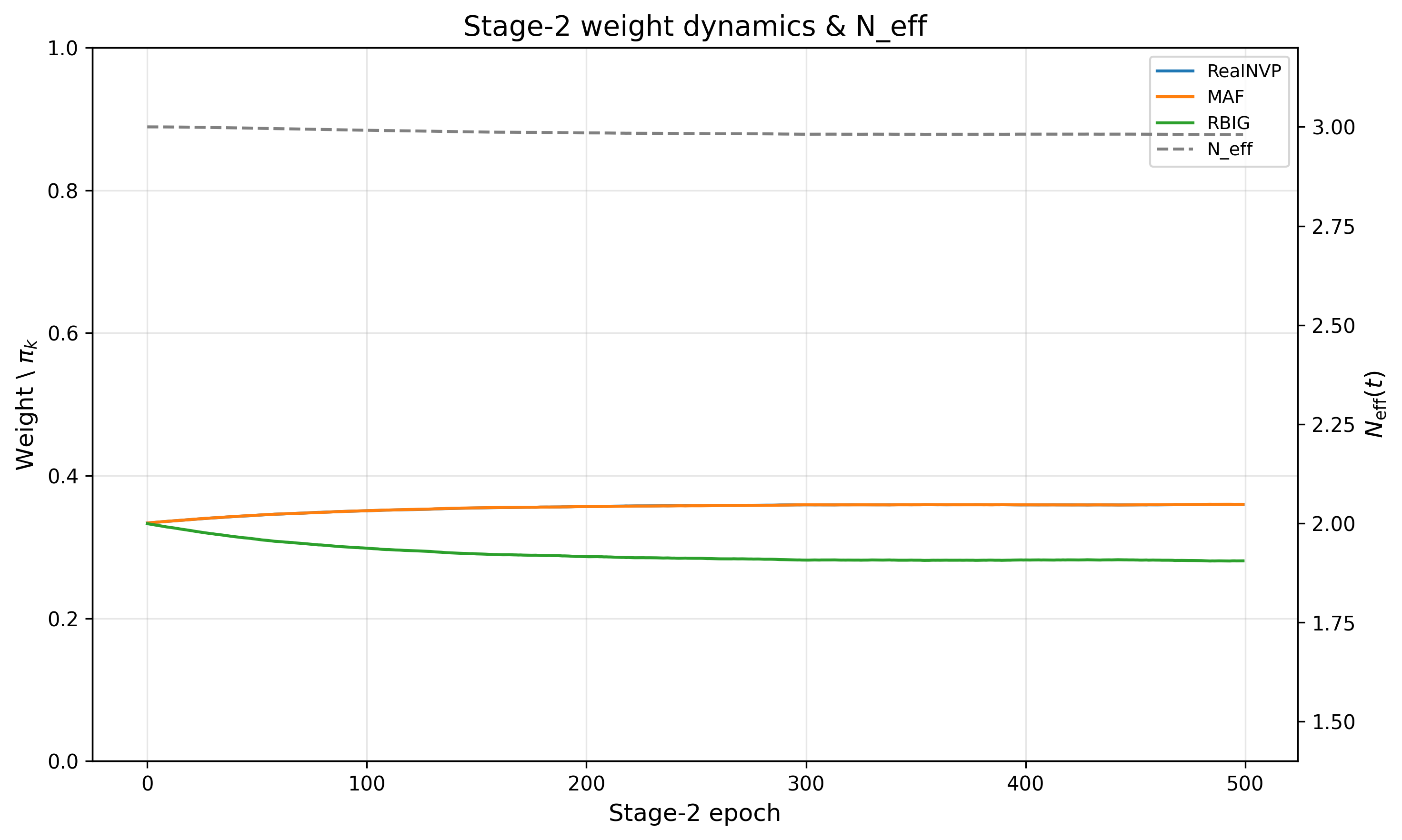}
    \caption{Multimodal: weight dynamics \& $N_{\text{eff}}$}
  \end{subfigure}\hfill
  \begin{subfigure}[t]{0.48\linewidth}
    \centering
    \includegraphics[width=\linewidth]{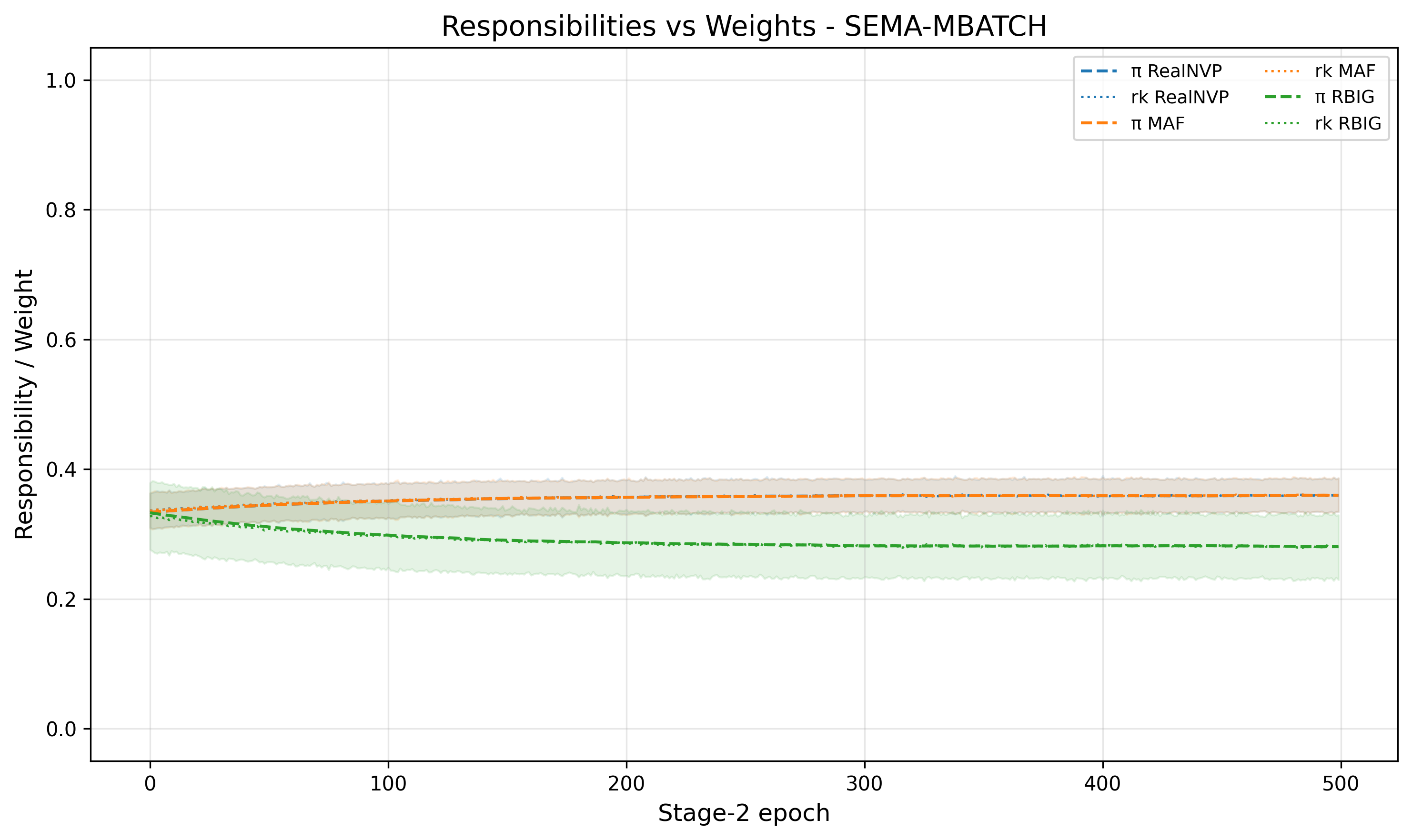}
    \caption{Multimodal: responsibilities vs.\ weights}
  \end{subfigure}
  \vspace{0.6em}
  % Rings
  \begin{subfigure}[t]{0.48\linewidth}
    \centering
    \includegraphics[width=\linewidth]{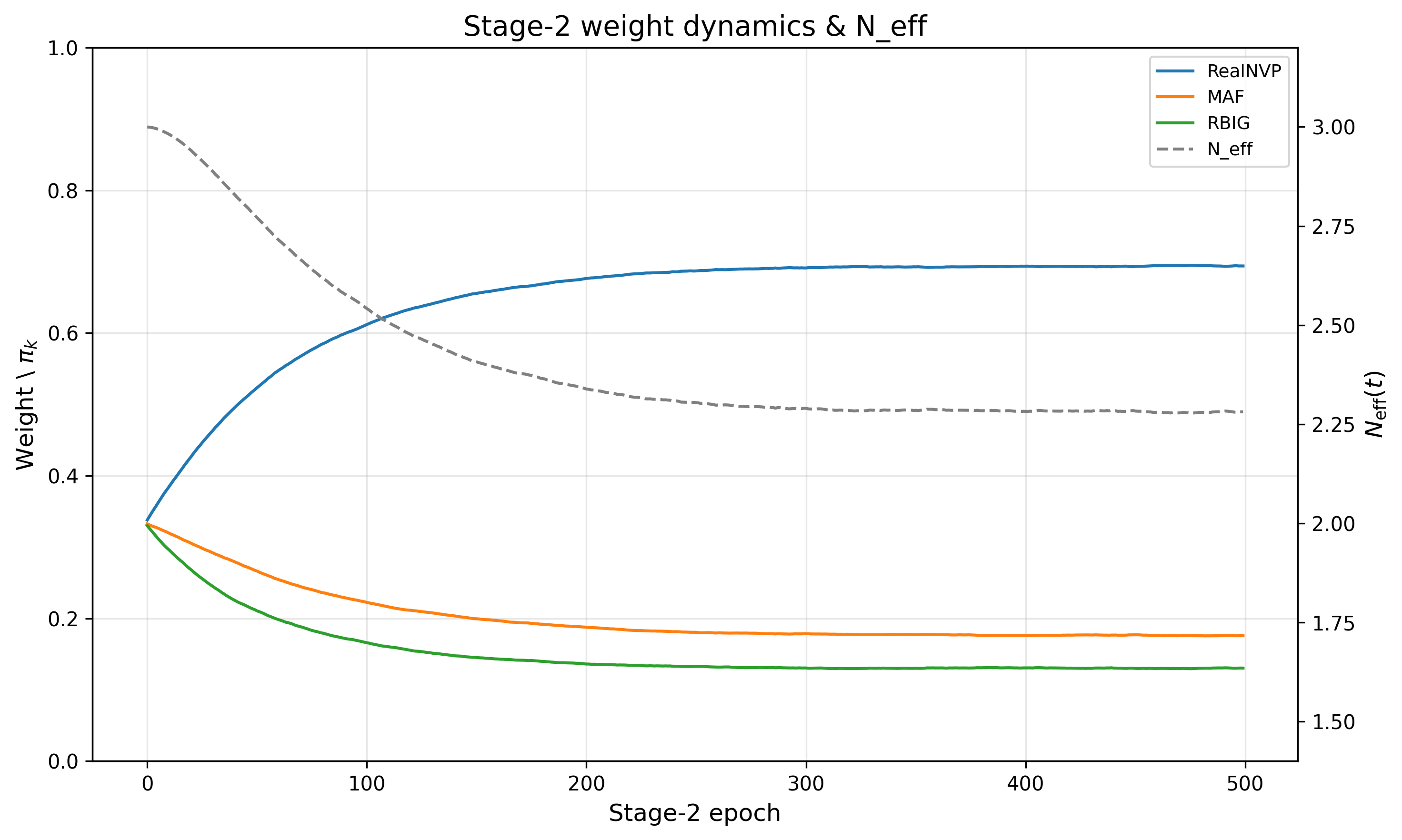}
    \caption{Rings: weight dynamics \& $N_{\text{eff}}$}
  \end{subfigure}\hfill
  \begin{subfigure}[t]{0.48\linewidth}
    \centering
    \includegraphics[width=\linewidth]{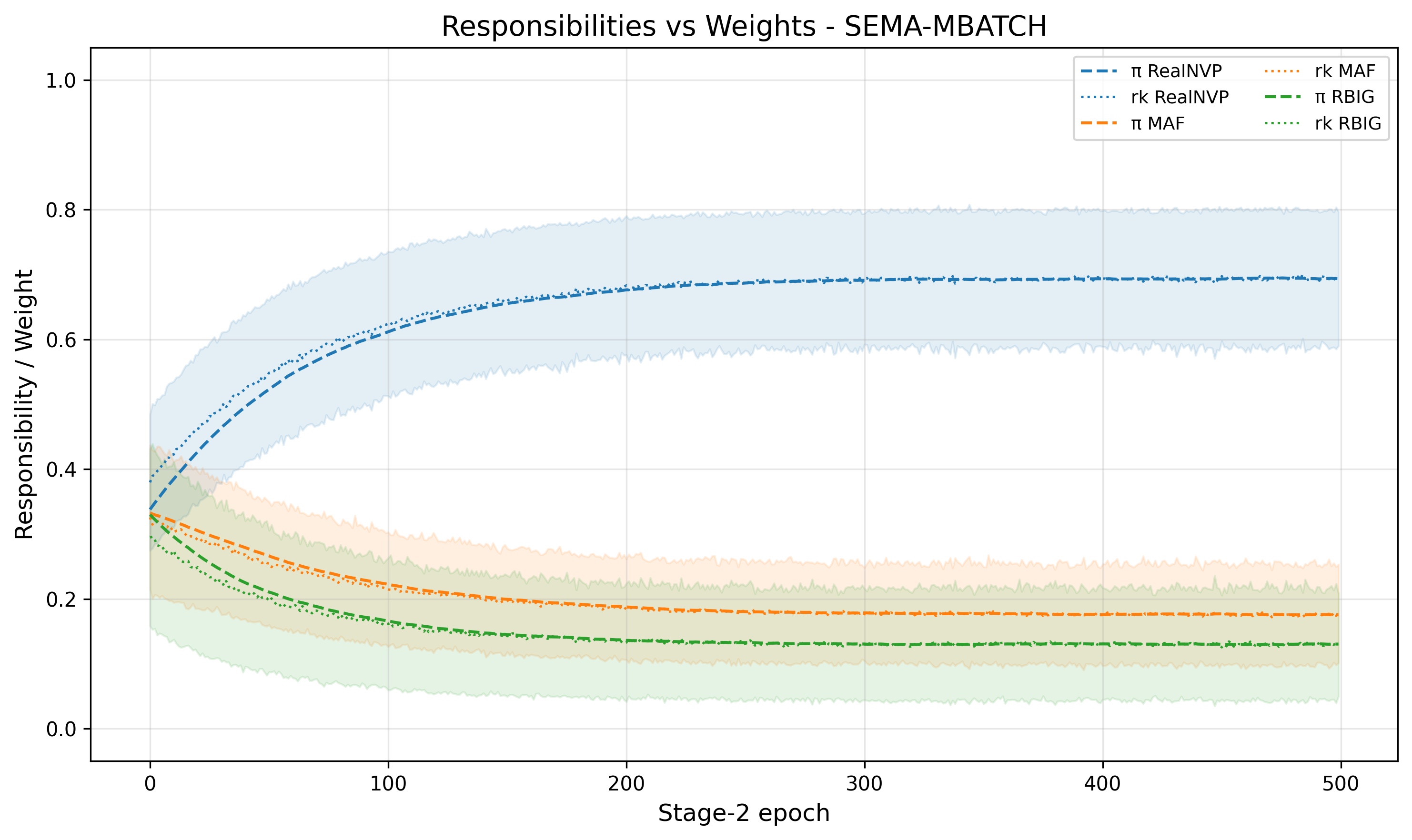}
    \caption{Rings: responsibilities vs.\ weights}
  \end{subfigure}
  \vspace{0.6em}
  % Real-GMM2
  \begin{subfigure}[t]{0.48\linewidth}
    \centering
    \includegraphics[width=\linewidth]{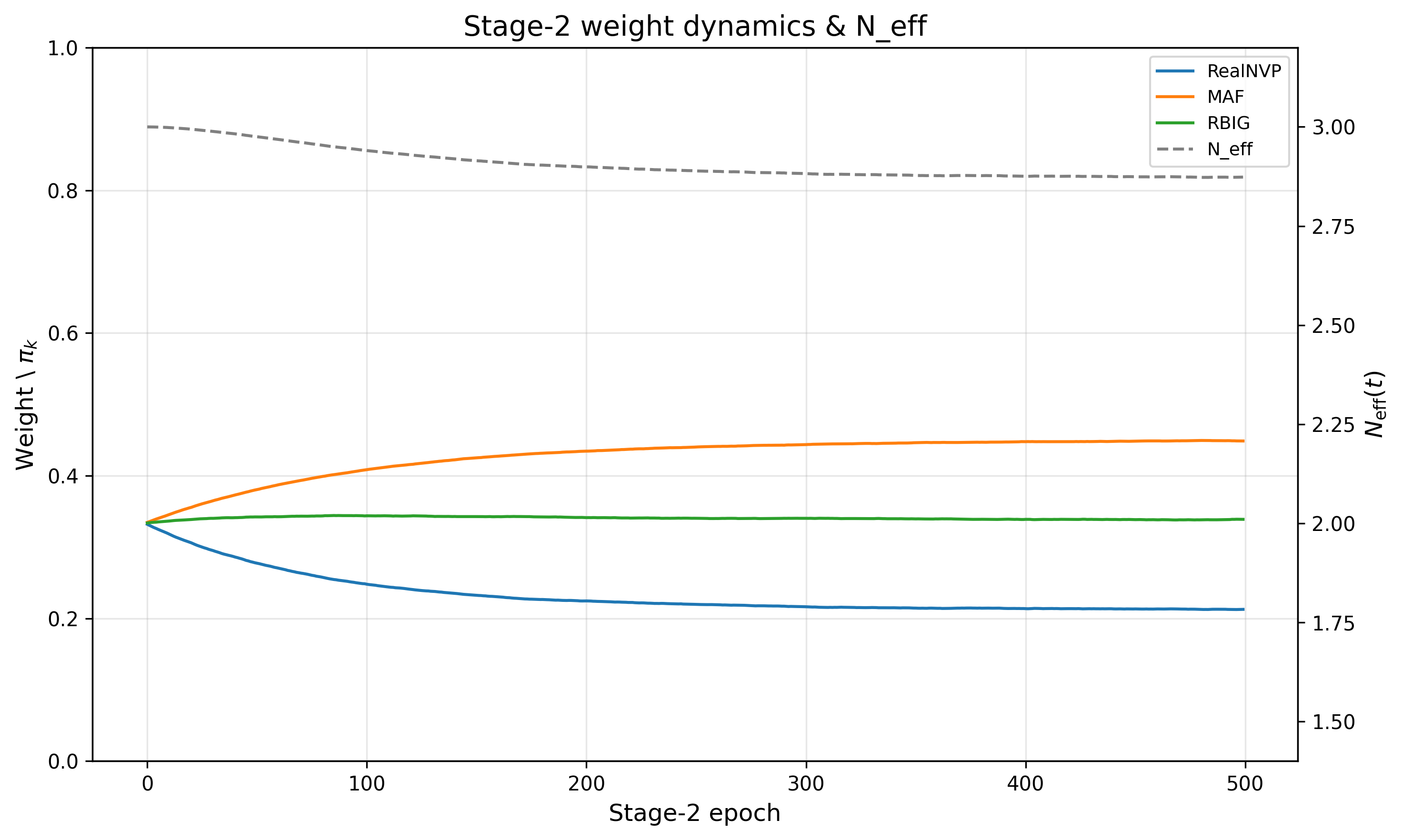}
    \caption{Real-GMM2: weight dynamics \& $N_{\text{eff}}$}
  \end{subfigure}\hfill
  \begin{subfigure}[t]{0.48\linewidth}
    \centering
    \includegraphics[width=\linewidth]{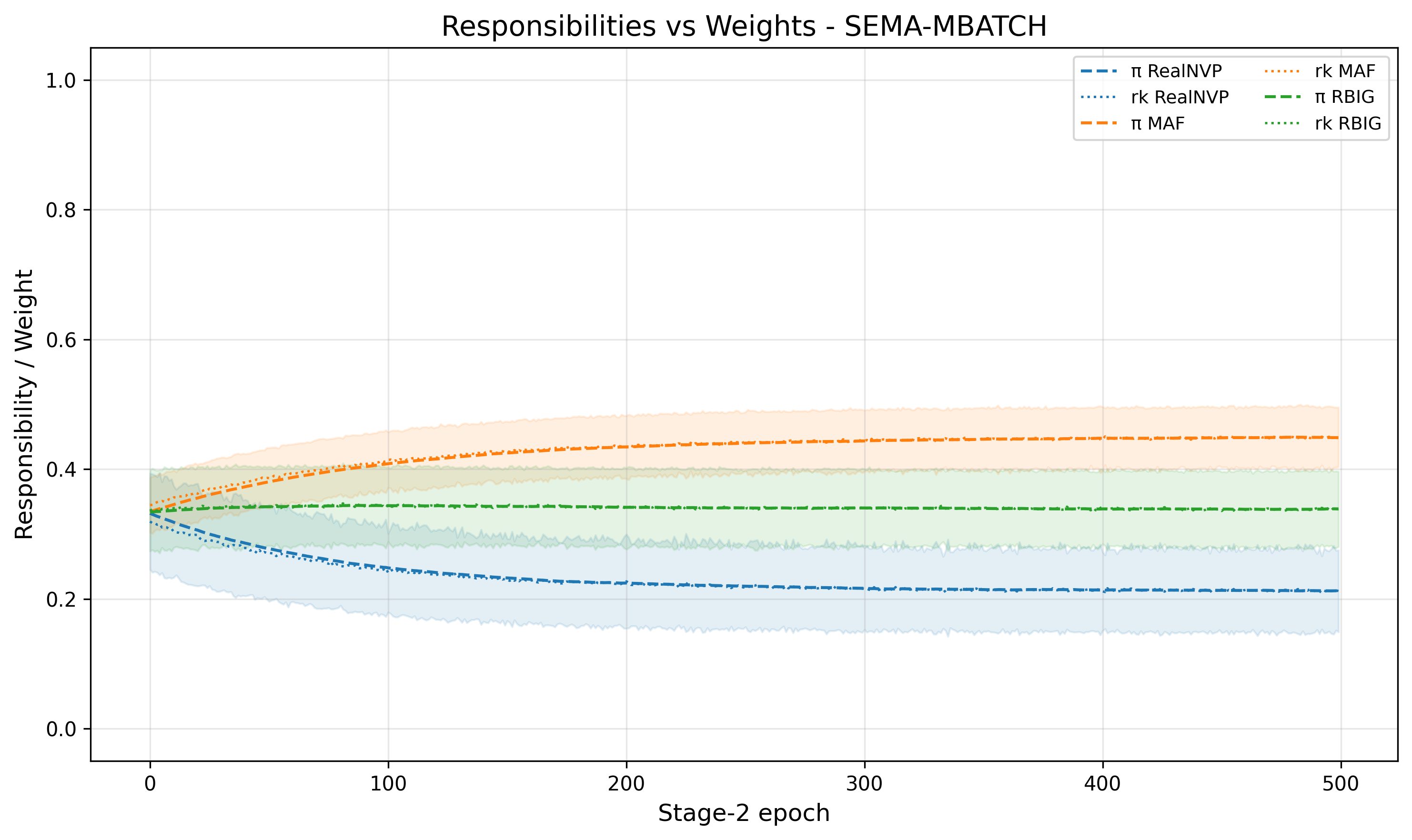}
    \caption{Real-GMM2: responsibilities vs.\ weights}
  \end{subfigure}
  \caption{Stage-2 gating dynamics of \textbf{AMF\mbox{-}VI\mbox{-}sEMA} across three representative targets.
  \textbf{Left} panels plot global mixture weights $\pi_k^{(t)}$ (solid lines) and the effective number of experts $N_{\text{eff}}(t){=}\exp(H(\pi^{(t)}))$ (right axis, dashed).
  \textbf{Right} panels compare per-epoch average responsibilities $\bar{r}_k^{(t)}$ (dotted) with the contemporaneous global weights $\pi_k^{(t)}$ (dashed), with shaded bands showing responsibility variance across batches.
  \emph{Multimodal-5} exhibits a nearly stationary regime: weights remain close to uniform ($\approx\!1/3$ each) with $N_{\text{eff}}\!\approx\!3.0$ throughout and negligible responsibility variance.
  \emph{Rings} shows a clear damped-convergence regime: \textsc{RealNVP} rises to $\approx\!0.70$ while \textsc{MAF} and \textsc{RBIG} decrease, with $N_{\text{eff}}$ settling at $\approx\!2.25$ and wide but stabilising responsibility bands.
  \emph{Real-GMM2} follows a slow, smooth trajectory: \textsc{MAF} gradually gains mass to $\approx\!0.45$, \textsc{RBIG} stabilises near $0.34$, \textsc{RealNVP} gently declines to $\approx\!0.21$, and $N_{\text{eff}}$ remains high ($\approx\!3.0$), indicating a near-stationary but weakly differentiated fixed point.}
  \label{fig:gate_dynamics_three}
\end{figure*}

Figure~\ref{fig:gate_dynamics_three} illustrates three characteristic regimes of the
sEMA gating dynamics, using Multimodal, Rings, and Real-GMM2 as representative targets.

\paragraph{Near-stationary regime (Multimodal).}
Figures~\ref{fig:gate_dynamics_three}(a)--(b) show the near-stationary regime.
In panel~(a), all three weights remain close to uniform throughout Stage~2: RealNVP and
MAF hold near ${\approx}0.35$--$0.36$ while RBIG declines gently to ${\approx}0.27$,
and $N_{\text{eff}}(t)$ stays flat at ${\approx}3.0$ from the very first epoch. There
is no transient and no meaningful drift, indicating the gate reaches a fixed point almost
immediately and all three experts contribute persistently. Panel~(b) confirms this:
the per-epoch responsibilities (dotted) closely track the global weights (dashed) with
narrow shaded bands throughout, reflecting negligible per-batch variance. This regime
requires no hyperparameter tuning; the defaults
$(\tau\!\approx\!1.1,\,\alpha\!\approx\!0.9,\,M\!=\!2)$ already provide smooth,
stable convergence.

\paragraph{Damped-convergence regime (Rings).}
Figures~\ref{fig:gate_dynamics_three}(c)--(d) show a markedly different picture.
Panel~(c) reveals rapid, monotonic specialisation: \textsc{RealNVP} rises from
${\approx}0.33$ to ${\approx}0.70$ within the first 200 epochs while \textsc{MAF} and
\textsc{RBIG} decline to ${\approx}0.18$ and ${\approx}0.14$ respectively, and
$N_{\text{eff}}(t)$ falls monotonically from ${\approx}3.0$ to ${\approx}2.25$. The
gate is actively concentrating mass on the best-suited expert for this annular geometry,
with no sign of convergence reversal. Panel~(d) explains this behaviour: the
per-batch responsibilities for \textsc{RealNVP} (blue dotted) exhibit very wide shaded
bands spanning roughly $0.2$--$0.8$, while the EMA weights (dashed) follow a much
smoother trajectory that tracks the mean responsibility signal. This large
responsibility variance is a signature of high per-batch likelihood variability on a
highly non-convex target; the EMA smoothing absorbs this noise and prevents individual
batches from causing large weight oscillations. In this regime, increasing $M$ or
$\alpha$ would further reduce responsibility variance and accelerate convergence to the
fixed point without changing the final weight allocation.

\paragraph{Intermediate regime (Real-GMM2).}
Figures~\ref{fig:gate_dynamics_three}(e)--(f) occupy a middle ground. Panel~(e) shows
slow, smooth drift: \textsc{MAF} rises gradually to ${\approx}0.45$, \textsc{RBIG}
stabilises near $0.34$, and \textsc{RealNVP} declines gently to ${\approx}0.21$, while
$N_{\text{eff}}(t)$ remains high at ${\approx}3.0$ and nearly flat throughout. The gate
is converging to a weakly differentiated fixed point rather than collapsing onto a single
expert. Panel~(f) confirms this: responsibilities (dotted) track the global weights
(dashed) with moderate but stable shaded bands, and all three experts maintain
well-separated yet smooth trajectories from the outset. This behaviour reflects a
target whose likelihood geometry is well-behaved and mildly informative about expert
competency, yielding a near-stationary fixed point with minimal update noise. It aligns
with the strong quantitative results for Real-GMM2 in Table~\ref{tab:results} (best
$W_2$ of $0.013$ and competitive NLL of $-1.619$), and suggests that for such targets
the Stage-2 budget ($M$ or epochs) could be reduced without harming final performance.

Taken together, the three cases confirm that sEMA produces qualitatively distinct but
well-behaved gating dynamics across different posterior geometries. \emph{Multimodal}
exemplifies the near-stationary regime, where all experts contribute persistently and
the gate requires no active smoothing. \emph{Rings} exemplifies damped convergence,
where rapid specialisation is accompanied by wide responsibility bands and the EMA
smoothing is doing the most work. \emph{Real-GMM2} occupies an intermediate regime of
slow, smooth drift toward a weakly differentiated fixed point. Across all three regimes
$N_{\text{eff}}(t){>}1$ throughout, confirming the absence of expert collapse. The key
practical implication is that wide responsibility bands (as in \emph{Rings}) are the
diagnostic signal for potential instability, and can be mitigated by increasing $M$ or
$\alpha$ without altering the final fixed point.

\begin{figure*}[ht!]
    \centering
    \begin{subfigure}[b]{0.98\textwidth}
        \centering
        \includegraphics[width=\linewidth]{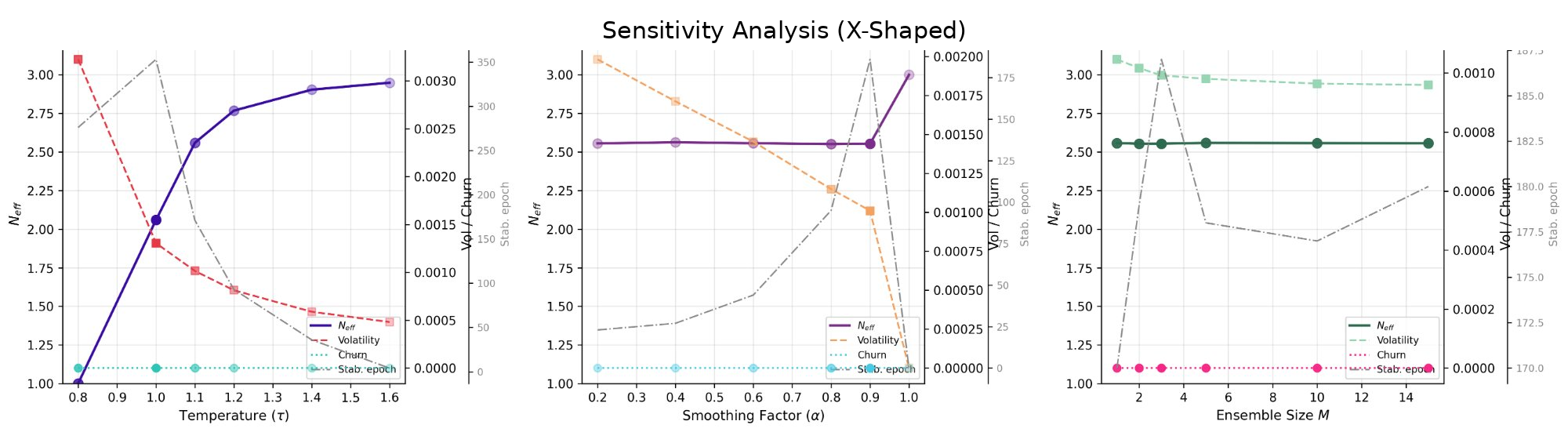}
        \caption{\texttt{X-shaped}}
    \end{subfigure}
    \hfill
    \begin{subfigure}[b]{0.98\textwidth}
        \centering
        \includegraphics[width=\linewidth]{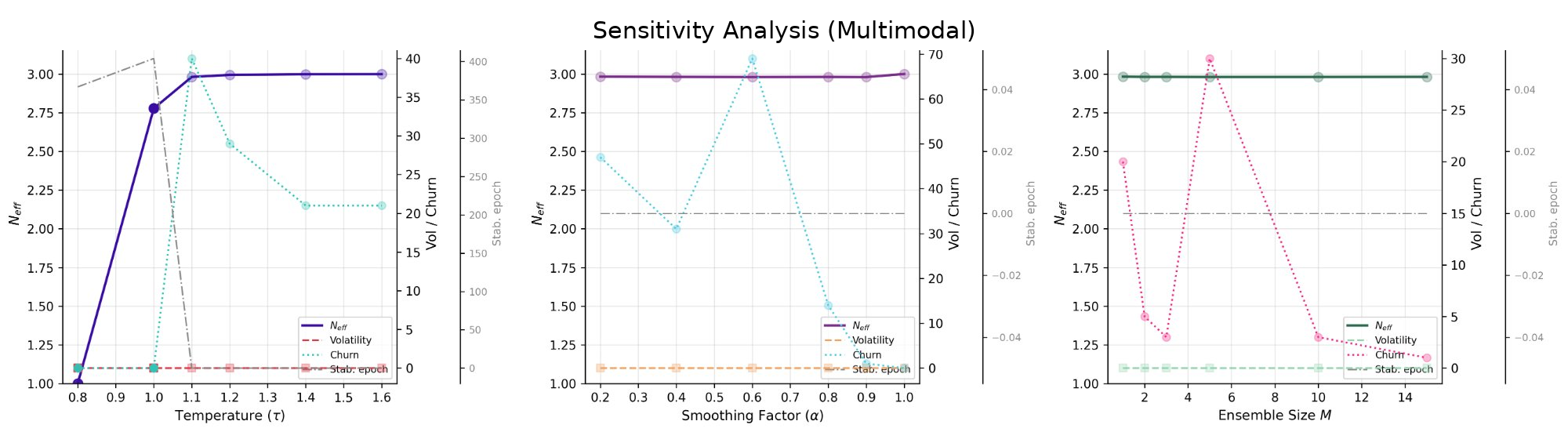}
        \caption{\texttt{Multimodal}}
    \end{subfigure}
    \caption{
        Sensitivity of \textit{AMF\mbox{-}VI\mbox{-}sEMA} to its Stage-2 hyperparameters
        $\tau$ (temperature), $\alpha$ (EMA smoothing), and $M$ (ensemble batch size)
        across two representative datasets (defaults: $\tau{=}1.1$, $\alpha{=}0.9$, $M{=}2$).
        Each panel reports the final effective expert count $N_{\text{eff}}{=}\exp(H(\pi))$ (solid),
        weight volatility $V$ (dashed), gate churn $C$ (dotted), and stabilisation epoch (dash-dot).
        On \texttt{X-shaped}, $\tau$ drives a smooth damped transition from collapse ($N_{\text{eff}}{\approx}1$
        at $\tau{=}0.8$) to near-uniform participation ($N_{\text{eff}}{\approx}2.95$ at $\tau{=}1.6$)
        with no churn. On \texttt{Multimodal}, the transition is sharper: collapse at $\tau{=}0.8$
        gives way abruptly to oscillatory behaviour ($C{>}0$) for $\tau{\geq}1.1$.
        Both $\alpha$ and $M$ have negligible effect on $N_{\text{eff}}$; only the degenerate
        $\alpha{=}1.0$ boundary forces exact uniform weights, confirming the EMA as the active
        diversity-preserving mechanism.
    }
    \label{fig:sensitivity}
\end{figure*}

\subsection{Sensitivity of sEMA to $\tau$, $\alpha$, and $M$}\label{sec:52}

To examine how the sEMA gate responds to its Stage-2 hyperparameters, we
systematically varied the temperature~$\tau$, smoothing factor~$\alpha$, and
ensemble size~$M$, and quantified three key behaviours:
\begin{enumerate}[label=(\roman*)]
  \item The effective number of active experts
    $N_{\text{eff}} = \exp(H(\pi_T))$
  \item Weight volatility
    $V = \tfrac{1}{T-1}\sum_t \lVert \pi_{t+1} - \pi_t \rVert_1$
  \item Churn $C$, counting the number of changes in the dominant expert
    over time.
\end{enumerate}

\paragraph{Temperature $\tau$.}
On \texttt{X-Shaped} (Figure~\ref{fig:sensitivity}a, left panel), increasing
$\tau$ progressively softens the gating distribution: $N_{\text{eff}}$ rises
monotonically from collapse (${\approx}1.0$ at $\tau{=}0.8$, stab.\ epoch
$276$) to ${\approx}2.95$ at $\tau{=}1.6$, while volatility declines steadily
and churn remains zero throughout, indicating a smooth damped regime across
the full sweep. On \texttt{Multimodal} (Figure~\ref{fig:sensitivity}b, left
panel), the transition is sharper: $\tau{=}0.8$ collapses the gate to a single
expert ($N_{\text{eff}}{\approx}1.0$, stab.\ epoch $367$), $\tau{=}1.0$
recovers broad participation ($N_{\text{eff}}{=}2.78$, stab.\ epoch $404$),
and $\tau{\geq}1.1$ saturates near the maximum ($N_{\text{eff}}{\approx}3.0$)
with low but non-zero churn (e.g.\ $C{=}40$ at $\tau{=}1.1$, declining to
$C{=}21$ at $\tau{=}1.6$), entering the oscillatory regime. Across both
datasets, $\tau$ is the dominant control parameter governing expert diversity.

\paragraph{Smoothing factor $\alpha$.}
The smoothing coefficient $\alpha$ has negligible effect on $N_{\text{eff}}$
across $\alpha{\in}[0.2,0.9]$ for both datasets: $N_{\text{eff}}$ remains
flat at ${\approx}2.55$ for \texttt{X-Shaped} and ${\approx}2.98$ for
\texttt{Multimodal} (Figures~\ref{fig:sensitivity}a--b, centre panels).
Volatility decreases monotonically with increasing $\alpha$ on both datasets,
confirming that higher momentum smooths weight trajectories. Only $\alpha{=}1.0$
(no EMA) forces exact uniform weights ($N_{\text{eff}}{=}3.0$, $V{=}0$,
$C{=}0$) on both datasets, confirming that the EMA update is the active
diversity-preserving mechanism: without it, all data-driven adaptation is lost
at the cost of higher NLL.

\paragraph{Ensemble size $M$.}
The ensemble size $M$ is similarly inert with respect to $N_{\text{eff}}$ and
NLL, which vary by less than $10^{-4}$ from $M{=}1$ to $M{=}15$ on both
datasets (Figures~\ref{fig:sensitivity}a--b, right panels). However, churn
reveals a meaningful distinction: on \texttt{Multimodal}, churn drops sharply
from $C{=}20$ at $M{=}1$ to $C{=}5$ at $M{=}2$, with further reductions
($C{=}3$ at $M{=}3$) yielding diminishing returns. On \texttt{X-Shaped},
churn is zero for all $M$, confirming that the damped regime is entirely
insensitive to batch count. Together, these results justify $M{=}2$ as the
default: it eliminates the instability present at $M{=}1$ in multimodal
settings at minimal additional cost, while producing effectively the same
final weights and NLL as larger $M$.

\paragraph{Summary.}
No parameter setting induces pathological behaviour beyond the known collapse
at $\tau{=}0.8$, and NLL remains stable to within ${\sim}0.002$ across all
non-degenerate configurations. The temperature $\tau$ governs the sharpness
of the softmax gating and thus expert diversity; $\alpha$ modulates weight
volatility and churn without affecting $N_{\text{eff}}$ except at the
degenerate boundary $\alpha{=}1.0$; and $M$ provides churn reduction in
oscillatory regimes with no effect on final weight allocation. Taken together,
these results confirm that sEMA provides a robust and interpretable gating
mechanism with a well-defined operating range that is largely insensitive to
hyperparameter choice within the ranges tested.

\subsection{Robustness and Generalization}
\label{sec:53}

\begin{figure}[t]
    \centering
    \begin{subfigure}{0.98\linewidth}
        \centering
        \includegraphics[width=\linewidth]{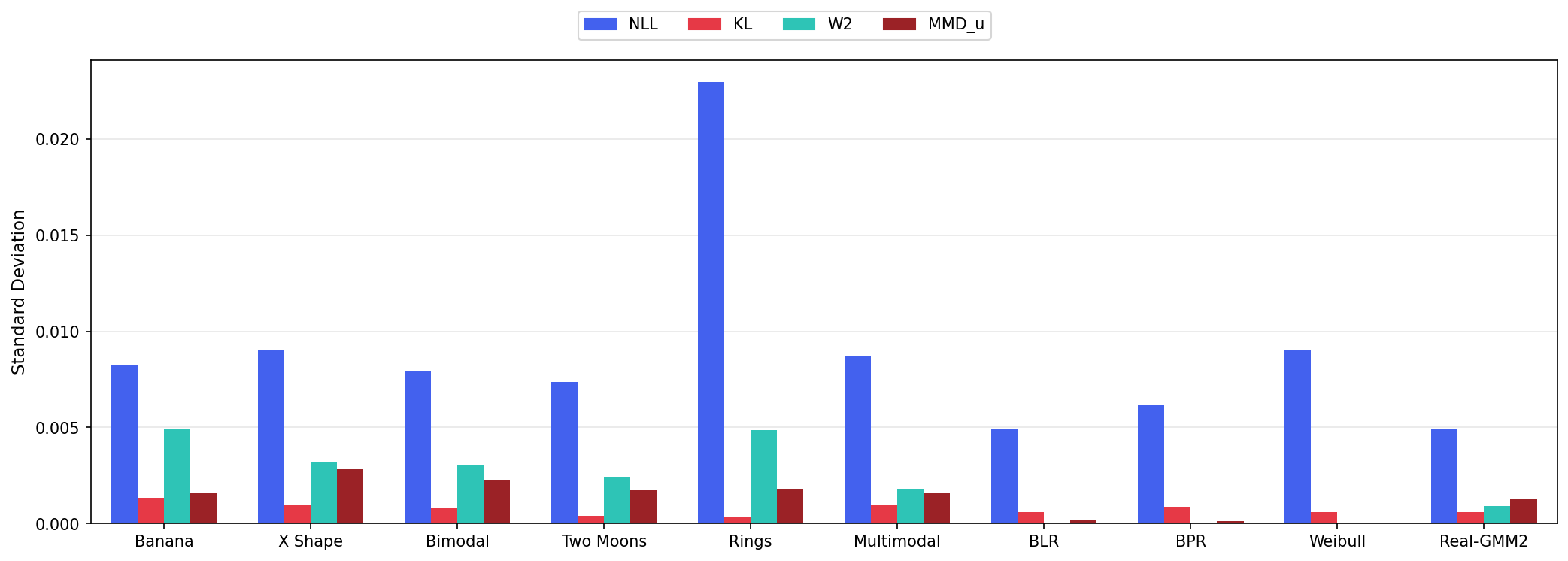}
        \caption{Standard deviation across random seeds.}
        \label{fig:robustness_std}
    \end{subfigure}
    \hfill
    \begin{subfigure}{0.98\linewidth}
        \centering
        \includegraphics[width=\linewidth]{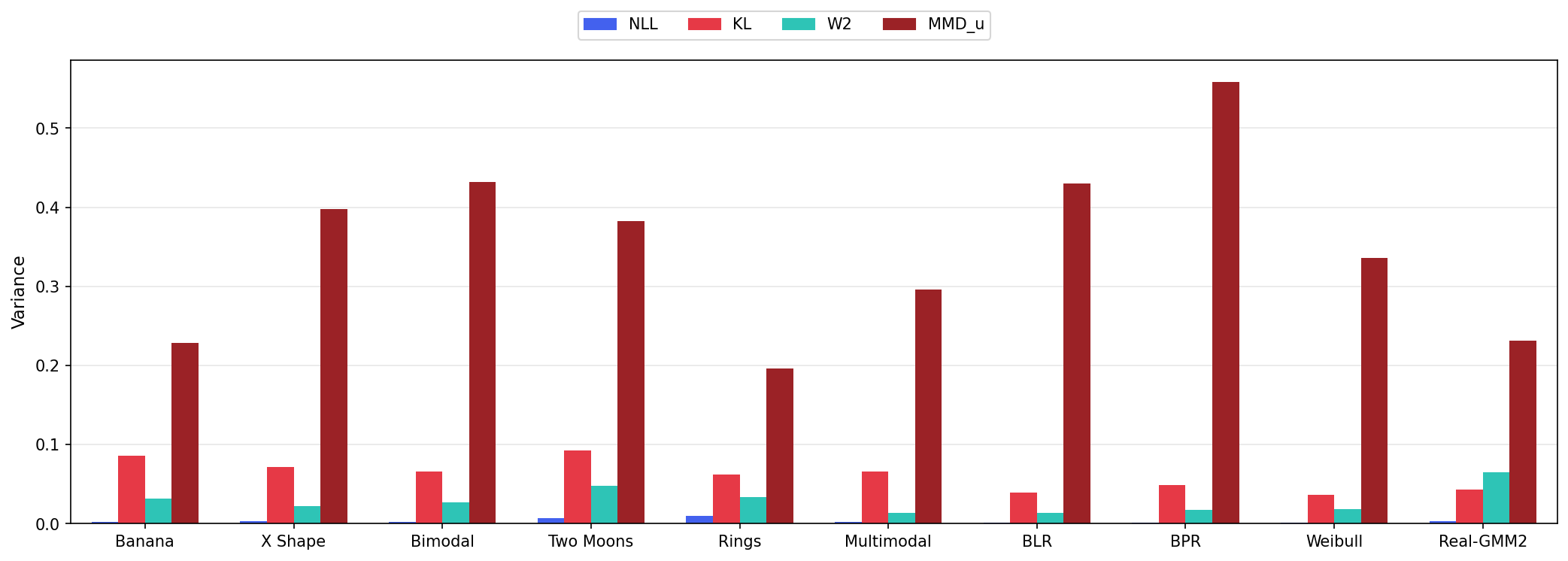}
        \caption{Coefficient of variation (CV) across seeds.}
        \label{fig:robustness_cv}
    \end{subfigure}
    \caption{
    Cross-seed robustness of \emph{AMF\mbox{-}VI\mbox{-}sEMA} across all ten datasets.
    Standard deviations (top) are small across all metrics and datasets: NLL std remains
    ${\leq}0.023$ (largest on \emph{Rings}), while KL, $W_2$, and MMD stds are near zero. The coefficient of variation (CV, bottom) indicates that the unbiased MMD exhibits higher relative variability (up to ${\approx}0.56$ on BPR) due to its near-zero absolute magnitude, whereas NLL, KL, and $W_2$ maintain consistently low relative dispersion across datasets.%The CV (bottom) reveals that the unbiased MMD exhibits higher relative variability (up to ${\approx}0.56$ on BPR) owing to its near-zero absolute magnitude, whereas NLL, KL, and $W_2$ CVs remain low and stable throughout.
    Together, these results confirm that \emph{AMF\mbox{-}VI\mbox{-}sEMA} is
    reproducible across random initialisations and sampling noise.
    }
    \label{fig:robustness_main}
\end{figure}

Figure~\ref{fig:robustness_main} summarises the cross-seed variability of
\emph{AMF\mbox{-}VI\mbox{-}sEMA} over ten datasets spanning unimodal, structured,
multimodal, and real posteriors, evaluated across 10 independent seeds (2025--2034). NLL standard deviations are uniformly small: ${\leq}0.009$ on all datasets except \emph{Rings} ($0.023$), which reflects its sharper likelihood landscape rather than instability. KL, $W_2$, and absolute MMD stds are near zero throughout (${\leq}0.005$), confirming tight reproducibility on transport and discrepancy metrics. The CV plot reveals that the elevated relative variability in MMD (up to $0.56$ on BPR) is an artefact of near-zero absolute MMD values rather than genuine instability, as NLL, KL, and $W_2$ CVs remain low across all datasets. Together, these results confirm that the sEMA gating mechanism is robust to random initialisation and sampling noise, with no dataset exhibiting pathological variance in any primary metric.

For well-behaved densities such as \texttt{X-Shaped} (NLL std $0.009$), \texttt{Banana} ($0.008$), and \texttt{Weibull} ($0.009$), dispersion of NLL and $\mathrm{KL}(p\|q)$ is minimal, illustrating that the sequential mixture updates converge to similar solutions across runs. Even for structured multimodal settings (\texttt{Bimodal} NLL std $0.008$, \texttt{Multimodal} $0.009$, \texttt{Rings} $0.023$), $W_2$ and absolute MMD stds remain ${\leq}0.005$, demonstrating that sEMA prevents expert collapse and maintains a balanced allocation of responsibilities. On real targets, variability is even lower: BLR and Real-GMM2 both achieve NLL std $0.005$, with $W_2$ std at or below $0.001$. The elevated
MMD coefficient of variation on real regression targets (BPR: $0.56$; BLR: $0.43$) is a consequence of near-zero absolute MMD values rather than genuine instability. The small KL variation across seeds (${\leq}0.001$ on most datasets) confirms that effective posterior coverage is reproducible even when component weights are adaptively updated through the stochastic EMA mechanism.

The highest NLL variability is observed for \texttt{Rings} (std $0.023$), whose sharp annular geometry makes likelihood estimates more sensitive to initialisation differences. The highest relative (CV) variability appears in the KL of \texttt{Two Moons} (${\approx}0.09$) and the MMD of real targets (BPR: $0.56$; BLR: $0.43$), the latter driven by near-zero absolute MMD values rather than genuine instability. Nonetheless, absolute magnitudes of both std and CV remain bounded across all datasets and metrics, confirming that the EMA-based weight updates effectively regularise expert transitions and prevent sensitivity to initialisation from propagating into metric-level instability.

Overall, these results confirm that the sEMA mechanism introduces a stabilising temporal coupling into the mixture weight updates: by smoothing expert weight trajectories via exponential moving averages, it promotes consistent training dynamics across all ten datasets and 10 independent seeds. NLL stds remain ${\leq}0.023$ everywhere, and KL/$W_2$ stds are an order of magnitude smaller, indicating that the posterior approximation quality is highly reproducible regardless of random initialisation, mini-batch ordering, or sampling noise. The apparent large CV values for unbiased MMD on real targets are a measurement artefact of near-zero denominators, not genuine instability. This across-seed robustness is essential for practical deployment of approximate inference methods, where reproducibility under varying stochastic conditions is a key indicator of reliability.

\subsection{Summary of Findings}
\label{sec:summary}

Across all experimental evaluations, the results consistently demonstrate that the proposed \emph{AMF\mbox{-}VI\mbox{-}sEMA} framework provides a stable and interpretable mechanism for adaptive mixture inference. The gate dynamics analysis (Sec.~\ref{sec:51}) revealed that the stochastic EMA effectively moderates expert specialisation, producing smooth responsibility transitions and reducing volatile weight updates. The sensitivity experiments (Sec.~\ref{sec:52}) showed that moderate variations of the hyperparameters $\tau$, $\alpha$, and $M$ exert limited influence on the effective number of active experts, confirming that the model's behaviour is largely invariant to tuning within a broad operational range. Finally, the robustness study (Sec.~\ref{sec:53}) established that performance variability across random seeds remains remarkably low (particularly for NLL and $\mathrm{KL}(p\|q)$), highlighting the reliability and reproducibility of the learned posteriors.

Taken together, these results indicate that \emph{sEMA} introduces a principled temporal regularisation effect that stabilises both optimisation and inference. 
It promotes consistent expert activation, mitigates stochastic drift, and preserves multimodal structure even under non-convex and noisy conditions. 
This combination of interpretability, robustness, and parameter insensitivity underlines the practical utility of \emph{AMF–VI–sEMA} as a general-purpose variational inference framework for complex posterior landscapes.

\section{Conclusion}
\label{sec:conclusion}

We presented \emph{AMF\mbox{-}VI\mbox{-}sEMA}, a temporally regularised extension of the Adaptive Mixture of Flows variational inference framework. Building upon the original two-stage formulation, in which heterogeneous flow experts (\textsc{MAF}, \textsc{RealNVP}, and \textsc{RBIG}) are first trained independently and subsequently combined through likelihood-based weighting, we introduced a stochastic exponential moving-average (sEMA) mechanism that smooths weight updates over time. This temporal regularisation stabilises the gating dynamics, mitigates volatile expert switching, and preserves mixture flexibility without requiring per-sample gating or joint optimisation of heterogeneous experts.

Across ten posterior benchmarks spanning six synthetic 2D families (\emph{Banana}, \emph{X-Shaped}, \emph{Bimodal}, \emph{Two-moons}, \emph{Rings}, and \emph{Multimodal}) together with four real/low-dimensional targets (BLR, BPR, Weibull, and Real-GMM2), \emph{AMF\mbox{-}VI\mbox{-}sEMA} achieved consistently competitive likelihood performance while maintaining strong transport and discrepancy metrics. Relative to the original \emph{AMF\mbox{-}VI}, the proposed sEMA weighting mechanism produced consistent improvements in NLL across all evaluated datasets while preserving stable KL, $W_2$, and MMD behaviour. Compared with individual flow baselines, the proposed framework demonstrated greater robustness across heterogeneous geometries, avoiding the severe transport failures observed in \textsc{ResFlow} and \textsc{EM\mbox{-}Mix} and reducing the geometry-specific regressions exhibited by methods such as \textsc{NICE}. Importantly, the learned global weights remained interpretable and non-collapsed, with effective expert counts $N_{\text{eff}}$ spanning $[1.41,\,2.99]$, indicating controlled geometry-aware specialisation rather than dominance by a single expert.

The extended analyses further highlighted the stability properties of the proposed framework. Gate-dynamics experiments revealed three characteristic regimes: near-stationary behaviour (\emph{Multimodal}, \emph{Real-GMM2}), damped convergence (\emph{Rings}), and intermediate slow-drift dynamics. Sensitivity studies showed stable behaviour across a broad range of $\tau$, $\alpha$, and $M$, with the temperature parameter $\tau$ acting as the primary controller of expert diversity, while ensemble size $M$ had minimal impact beyond $M{=}2$. Cross-seed evaluations over ten independent runs demonstrated low NLL variance (${\leq}0.023$) together with near-zero KL and $W_2$ variability across all datasets, confirming robustness to random initialisation and stochastic sampling effects.

Despite these encouraging results, several limitations remain. The current framework employs global mixture weights and has not yet been evaluated on very high-dimensional or strongly constrained posterior families. Moreover, although the proposed gating mechanism improves stability and interpretability, it does not explicitly model input-dependent expert allocation. Future work will therefore explore conditional or amortised gating strategies, adaptive expert selection, and extensions to higher-dimensional Bayesian inference settings.

Overall, \emph{AMF\mbox{-}VI\mbox{-}sEMA} demonstrates that temporally regularised global weighting provides a simple, interpretable, and computationally efficient mechanism for stabilising heterogeneous flow mixtures. By combining adaptive expert allocation with architecture-agnostic training, the framework offers a practical and robust approach for modelling complex, non-convex, and multimodal posterior geometries.

\appendix
\section{Normalising Flows and Information-Theoretic Background}
\label{app:flows_and_metrics}

\subsection{Normalising Flows and Likelihood Computation}

A normalising flow constructs a complex density $p_X(x)$ by transforming a simple base distribution $p_Z(z)$ through an invertible mapping
\[
f_\theta : \mathbb{R}^d \rightarrow \mathbb{R}^d .
\]
Using the change-of-variables formula, the transformed density can be written as
\begin{equation}
p_X(x)
=
p_Z(f_\theta^{-1}(x))
\left|
\det J_{f_\theta^{-1}}(x)
\right|
=
p_Z(z)
\left|
\det J_{f_\theta}(z)
\right|^{-1},
\end{equation}
where $J_{f_\theta}$ denotes the Jacobian of the transformation and
\[
z = f_\theta^{-1}(x).
\]
For a composition of $K$ invertible transformations
\[
f_\theta = f_K \circ \cdots \circ f_1,
\]
the log-density becomes
\begin{equation}
\log p_X(x)
=
\log p_Z(z_K)
-
\sum_{k=1}^{K}
\log
\left|
\det J_{f_k}(z_{k-1})
\right|.
\end{equation}
This formulation enables exact likelihood evaluation while progressively transforming a simple latent distribution into a highly expressive target density capable of representing multimodal, curved, and non-convex geometries.

\subsection{Maximum Likelihood and KL Divergence}

Normalising flows are typically trained through maximum likelihood estimation by minimising the empirical negative log-likelihood (NLL):
\begin{equation}
\mathcal{L}_{\text{NLL}}
=
-
\frac{1}{N}
\sum_{i=1}^{N}
\log p_X(x_i;\theta).
\end{equation}
This objective is equivalent to minimising the empirical cross-entropy between the data distribution $\hat{p}_{\text{data}}$ and the model distribution $p_X$:
\begin{equation}
\mathcal{L}_{\text{NLL}}
=
H(\hat{p}_{\text{data}}, p_X)
=
H(\hat{p}_{\text{data}})
+
D_{\text{KL}}(\hat{p}_{\text{data}} \| p_X).
\end{equation}
Since the entropy term
\[
H(\hat{p}_{\text{data}})
\]
does not depend on the model parameters $\theta$, maximising likelihood is equivalent to minimising the forward Kullback--Leibler divergence
\[
D_{\text{KL}}(p_{\text{data}} \| p_X).
\]

\subsection{Information-Theoretic Perspective}

Minimising the forward KL divergence penalises regions where
\[
p_{\text{data}}(x) > 0
\]
but
\[
p_X(x) \approx 0,
\]
thereby encouraging \emph{mode-covering} behaviour. In contrast, the reverse KL divergence commonly used in traditional variational inference,
\[
D_{\text{KL}}(q \| p),
\]
is generally \emph{mode-seeking}, often favouring sharp approximations that underestimate posterior support.

The proposed AMF--VI--sEMA framework inherits the forward-KL perspective through likelihood-based optimisation of each expert flow. Consequently, different experts are encouraged to capture complementary regions of probability mass, which naturally supports multimodal approximation and mitigates the risk of collapse onto a single dominant mode.

\section{Evaluation Metrics}
\label{app:metrics}

We evaluate mixture quality using four complementary metrics: negative log-likelihood (NLL), Kullback--Leibler (KL) divergence, Wasserstein-2 ($W_2$) distance, and maximum mean discrepancy (MMD). Together, these metrics characterise likelihood fidelity, distributional divergence, geometric alignment, and sample-level similarity.

\paragraph{Negative Log-Likelihood (NLL).}
The negative log-likelihood measures how well the learned density explains observed samples:
\begin{equation}
\text{NLL}
=
-
\frac{1}{N}
\sum_{i=1}^{N}
\log p_X(x_i),
\end{equation}
which is equivalent to the empirical cross-entropy
\[
H(p_{\text{data}}, p_X).
\]
Lower NLL values indicate better likelihood fit to the target distribution.

\paragraph{Kullback--Leibler Divergence.}
The KL divergence quantifies the discrepancy between two probability distributions:
\begin{equation}
D_{\text{KL}}(p \| q)
=
\int
p(x)
\log
\frac{p(x)}{q(x)}
\, dx.
\end{equation}
Within AMF--VI--sEMA, we employ the forward divergence
\[
D_{\text{KL}}(p_{\text{data}} \| p_X)
\]
to assess overall distributional fidelity and mode coverage.

\paragraph{Wasserstein-2 Distance.}
The Wasserstein-2 distance measures geometric displacement between distributions:
\begin{equation}
W_2(p, q)
=
\left(
\inf_{\gamma \in \Pi(p,q)}
\int
\|x-y\|^2
\, d\gamma(x,y)
\right)^{1/2},
\end{equation}
where $\Pi(p,q)$ denotes the set of all valid couplings between $p$ and $q$. Empirically, $W_2$ is approximated using discrete optimal transport with uniform sample weights. Unlike likelihood-based metrics, Wasserstein distance is sensitive to geometric support alignment and spatial displacement of probability mass.

\paragraph{Maximum Mean Discrepancy (MMD).}
Maximum mean discrepancy quantifies the difference between two probability distributions, $p$ and $q$, through distances between their kernel mean embeddings in a reproducing kernel Hilbert space (RKHS) $\mathcal{H}_k$:
\begin{align}
\text{MMD}^2(p,q)
&=
\sup_{\|f\|_{\mathcal{H}_k}\leq1}
\left|
\mathbb{E}_{x\sim p}[f(x)]
-
\mathbb{E}_{y\sim q}[f(y)]
\right|.
\end{align}
Applying the kernel trick yields the equivalent form
\begin{align}
\text{MMD}^2(p,q)
&=
\mathbb{E}_{x,x'\sim p}[k(x,x')]
+
\mathbb{E}_{y,y'\sim q}[k(y,y')]
-
2\mathbb{E}_{x\sim p,\,y\sim q}[k(x,y)],
\end{align}
where we employ a Gaussian radial basis function (RBF) kernel
\[
k(x,y)
=
\exp\!\left(
-\frac{\|x-y\|^2}{2\sigma^2}
\right).
\]
MMD provides a flexible non-parametric measure of sample similarity and is particularly useful for evaluating multimodal sample quality.

\begin{table}[h]
\centering
\caption{Summary of evaluation metrics. $^*$\textit{MMD is a true metric when the kernel is characteristic (e.g., Gaussian RBF).}}
\label{tab:appendix}
\begin{tabular}{lcccc}
\hline
\textbf{Property} & \textbf{NLL} & \textbf{KL} & \textbf{$W_2$} & \textbf{MMD} \\
\hline
Metric & No & No & Yes & Yes$^*$ \\
Symmetric & N/A & No & Yes & Yes \\
Mode coverage & Yes & Partial & Yes & Yes \\
Geometric awareness & No & No & Yes & Kernel-dependent \\
Sample complexity & $O(n)$ & $O(n)$ & $O(n^3)$ & $O(n^2)$ \\
\hline
\end{tabular}
\end{table}

\paragraph{Metric Summary.}
The complementary properties of the evaluation metrics are summarised in Table~\ref{tab:appendix}. Joint consideration of NLL, KL divergence, Wasserstein distance, and MMD enables assessment of likelihood quality, divergence behaviour, geometric fidelity, and sample-level agreement simultaneously, providing a comprehensive evaluation of the proposed AMF--VI--sEMA framework.

\vskip 0.2in
\bibliography{refs}

\end{document}